\documentclass[journal]{IEEEtran}
\usepackage[T1]{fontenc}
\usepackage{amsmath,amssymb,bm}
\usepackage{graphicx}
\usepackage{booktabs}
\usepackage{siunitx}
\usepackage{xcolor}
\usepackage{tikz}
\usetikzlibrary{arrows.meta,positioning,shapes.geometric,fit,calc,backgrounds}
\usepackage{algorithm}
\usepackage{algpseudocode}
\usepackage{cite}
\usepackage{microtype}
\usepackage{multirow}
\usepackage[hidelinks]{hyperref}

\newcommand{\NdesRuns}{12}
\newcommand{\ATEdesRange}{0.47--1.91~m}
\newcommand{\ATEalignedDesRange}{0.18--0.53~m}
\newcommand{\ATEodoDesRange}{4.2--19.0~m}
\newcommand{\PrecDesRange}{0.994--1.000}

\newcommand{\CommitsDes}{457}

\newcommand{\CovDesRange}{0.79--0.84}

\newcommand{\ATEalignedDegRange}{0.25--0.81~m}

\newcommand{\OdoDriftRange}{4--36~m}

\newcommand{\TopOneFinalIndoor}{0.91}
\newcommand{\TopOneEnergyIndoor}{0.88}
\newcommand{\SensorTopOneBest}{0.89}
\newcommand{\CaptureAlong}{\SI{\pm 0.20}{\meter}}
\newcommand{\CaptureLateral}{\SI{\pm 0.15}{\meter}}
\newcommand{\CaptureHeading}{\SI{\pm 25}{\degree}}

\newcommand{\CostTotal}{69}

\newcommand{\CostGraph}{41}

\newcommand{\LLRmid}{0.70}
\newcommand{\LLRslope}{40}
\newcommand{\Smin}{0.60}

\newcommand{\RealPulses}{1186}
\newcommand{\RealDist}{89}

\newcommand{\RealSimTrueMed}{0.65}
\newcommand{\RealSimFalsePnn}{0.645}

\newcommand{\RealMidEven}{0.576}

\newcommand{\RealMidOdd}{0.546}

\newcommand{\RealATESim}{0.77~(0.07)}

\newcommand{\RealATEEven}{0.67~(0.13)}
\newcommand{\RealATEalEven}{0.52~(0.11)}
\newcommand{\RealCovEven}{0.64~(0.10)}

\newcommand{\RealWrongLinksEven}{3.3~(2.9)}

\newcommand{\RealATEodo}{3.62~(0.46)}

\newcommand{\RealATEcliffEven}{3.18~(0.21)}

\newcommand{\RealATEcliffNoise}{0.66~(0.07)}

\newcommand{\RealRateHz}{7.1}
\newcommand{\RealBudgetMs}{141}
\newcommand{\RealFrontendMs}{4.8}
\newcommand{\RealSlamMs}{6.2}

\title{BatSLAM 2.0: Sequence-Verified Sonar Place Recognition in a Robust Pose Graph}
\author{Jan~Steckel%
\thanks{J. Steckel is with Cosys-Lab, Faculty of Applied Engineering, University of Antwerp, 2020 Antwerp, Belgium, and also with Flanders Make Strategic Research Centre, 3920 Lommel, Belgium (e-mail: jan.steckel@uantwerpen.be).}}

\begin{document}
\maketitle

\begin{abstract}
Echolocating bats can navigate dark and cluttered spaces using echolocation. Over a decade ago, BatSLAM showed that a robot with a biomimetic binaural sonar can build a topological map of the environment, by recognizing places from the received acoustic signals. Sonar place recognition, however, is ambiguous by nature: corridors produce nearly identical echo trains, and wrong loop closure can collapse the topological map. In this paper, we introduce BatSLAM 2.0, a novel sonar-only SLAM system built from three elements: an updated acoustic front-end, a sequence verifier that tracks and verifies loop closure candidates and a pose graph implemented on a high performance factor graph framework. The system was thoroughly evaluated both in simulated as well as real world recordings. In both cases, the BatSLAM2.0 algorithm shows the capability of robust topological map creation, countering map collapse, and robust scaling of map size.
\end{abstract}


\section{Introduction}
Bats are the prime example of an animal that navigates using sound alone. Indeed, echolocating bats find their way through caves, forests and buildings using the echoes of their own calls, often in complete darkness and at high speeds \cite{griffinListeningDarkAcoustic1958, ulanovskyWhatBatVoice2008}. From an engineering perspective, this ability is more than a biological curiosity as acoustic sensing keeps working in exactly those conditions in which optical sensors such as cameras and lidar sensors struggle (darkness, smoke, dust, fog and transparent or reflective surfaces). Furthermore, ultrasonic sensors themselves are cheap, small and robust and can perform high resolution imaging on complex environments \cite{steckelBroadband3DSonar2013, kerstensERTISFullyEmbedded2019}. 

Early work on sonar-based mapping modeled the environment as a set of geometric primitives (e.g. planes, edges and cilinders), which were tracked in an extended Kalman filter \cite{leonardMobileRobotLocalization1991, tardosRobustMappingLocalization2002}. This approach requires reliable extraction of such primitives from sparse and noisy range readings, which is not always possible in real-world scenarios.

In our earlier work we implemented a biologically inspired SLAM system called BatSLAM \cite{steckelBatSLAMSimultaneousLocalization2013}. In this algorithm we combined a biomimetic binaural sonar (where the receivers replicated the outer ears of a bat), with a topological model called RatSLAM. This RatSLAM model is a biologically inspired SLAM system based on a model of the rodent hippocampus \cite{milfordRatSLAMHippocampalModel2004, milfordMappingSuburbSingle2008} that creates topological maps of the environment that simultaneously has geometrically valid properties. Instead of extracting landmarks from the received signals, BatSLAM used the complete binaural cochleogram of each echo train as a local view, and recognized robot poses by comparing these views with a database of templates. Later work confirmed that bat-like sonar views indeed carry enough information to recognize real-world complex places \cite{vanderelstPlaceRecognitionUsing2016}. The weak point of such an approach, however, is ambiguity. Many places produce similar local view templates (ie, similar echo trains received by the sonar sensor). A typical example is the two parallel corridors with plain walls that produce nearly constant echo trains irrespective of the position of the robot in that corridor. This can cause false place matches to be injected into the map with full confidence, that subsequently pulls geometrically distinct places onto each other. In the SLAM literature, this failure mode is called map collapse, and it is one of the main reasons why loop closure is typically treated as the most dangerous step in the SLAM pipeline \cite{cadenaPastPresentFuture2016, lowryVisualPlaceRecognition2016}.

Typically, two families of techniques are used against false loop closures. On the one hand, the front-end can demand temporal consistency, and compare sequences of views rather than single views, as in SeqSLAM \cite{milfordSeqSLAMVisualRoutebased2012}. On the other hand, the back-end can treat loop closures as potentially being outliers (i.e. false loop closures), using robust cost functions, switchable constraints \cite{sunderhaufSwitchableConstraintsRobust2012}, dynamic covariance scaling \cite{agarwalRobustMapOptimization2013}, max-mixtures \cite{olsonInferenceNetworksMixtures2013}, clustering and consistency checks \cite{latifRobustLoopClosing2013, mangelsonPairwiseConsistentMeasurement2018} or graduated non-convexity methods \cite{yangGraduatedNonConvexityRobust2020}. Both families were developed mostly for SLAM systems relying on vision and lidar, but could fundamentally be applied to other modalities such as in-air sonar. The question of how well they carry over to the much more ambiguous sonar views has, to the best of our knowledge, not been answered so far.

In this paper, we introduce BatSLAM 2.0, a novel sonar-only SLAM system that revisits the original BatSLAM idea with a modern and robust SLAM back-end. The system replaces the pose cells and the experience map of BatSLAM with a factor-graph pose graph solved with iSAM2 \cite{kaessISAM2IncrementalSmoothing2012} in GTSAM \cite{dellaertFactorGraphsGTSAM2012}, and it keeps local view templates that are anchored to pose-graph nodes, so that each template has a position and an orientation that move with the optimized pose graph. 

Around this core, we contribute four elements. Firstly, an acoustic front-end that adds an explicit direction cue to the  cochleogram: the spectral shape of each echo, which carries the directivity of the ears and the emitter, is separated from the echo magnitude, to more explicitly encode the direction of origin of the reflection. Secondly, a sequence verifier that keeps competing recognition hypotheses optional, and commits to a decision only when it is the loop closure is unambiguously valid. The required evidence to accept a loop closure grows with the geometric correction such commit would force on the map, and plausibility is tested against the relative uncertainty between the matched places. 

Thirdly, we propose a link management system in the SLAM back-end, in which the links of every committed hypothesis form a group that can be withdrawn, audited and re-admitted, which allows false loop closures to be pruned efficiently. Fourthly, a simulation study of which information in a sonar view makes a place recognizable, and of the capture regions of individual view templates. Finally, we validate the system on a real recording of an embedded 3D sonar \cite{kerstensERTISFullyEmbedded2019}, whose microphone array we turned into two virtual ears by beamforming, with a lidar-based ground truth. A video of BatSLAM 2.0 in operation is available online.\footnote{\label{fn:video}\url{https://youtu.be/VzuYBWAR8kw}}

The rest of the paper is structured as follows. First, we give an overview of the BatSLAM 2.0 system in section \ref{sec:overview}. Next, we derive the acoustic front-end in section \ref{sec:frontend}, after which we describe the recognition and verification of places in section \ref{sec:recognition} and the robust pose graph in section \ref{sec:backend}. In section \ref{sec:setup}, we detail the simulation setup, and in section \ref{sec:results} we present the results, including ablations and a robustness analysis. In section \ref{sec:real}, we apply the system to a real recording. Finally, we discuss the limitations of our work in section \ref{sec:limitations} and conclude the paper in section \ref{sec:conclusion}.

\section{Overview of BatSLAM 2.0}
\label{sec:overview}
\begin{figure*}[t]
\centering
\begin{tikzpicture}[
  box/.style={draw=black!70, rounded corners=2pt, align=center, font=\footnotesize, text width=2.45cm, minimum height=#1},
  box/.default=1.75cm,
  arr/.style={-{Stealth[length=2mm]}, thick, black!70},
  lab/.style={font=\scriptsize, black!65, align=center}]
\node[box, fill=blue!6] (echo) at (0,0) {Binaural Echo\\ $s_L(t)$, $s_R(t)$};
\node[box, fill=blue!6] (fe) at (3.4,0) {Acoustic Front-End\\ Local view $V$:\\ energy $E$, shape $S$\\ (Section \ref{sec:frontend})};
\node[box, fill=blue!6] (tm) at (6.8,0) {Template Store\\ Five best matches\\ (Section \ref{sec:templates})};
\node[box, fill=orange!10] (sv) at (10.2,0) {Sequence Verifier\\ Hypotheses,\\ evidence, plausibility,\\ risk-scaled commit\\ (Section \ref{sec:verifier})};
\node[box, fill=green!8] (pg) at (13.6,0) {Pose Graph\\ (iSAM2)\\ Odometry and\\ weak links\\ (Section \ref{sec:backend})};
\node[box=1.0cm, fill=blue!6] (lay) at (6.8,-2.5) {Lay New Template\\ Anchored to node $i$};
\node[box=1.0cm, fill=green!8] (lm) at (13.6,-2.8) {Link Management\\ Length rule, residual\\ check, GNC audit};
\node[box=0.7cm, fill=white] (odo) at (14.2,2.3) {Odometry $u_i$};
\draw[arr] (echo) -- (fe);
\draw[arr] (fe) -- (tm);
\draw[arr] (tm) -- (sv);
\draw[arr] (sv) -- (pg);
\draw[arr] (odo.south) -- (odo.south |- pg.north);
\draw[arr] (pg) -- (lm);
\draw[arr, dashed] ($(pg.north)+(-0.6,0)$) -- ++(0,0.5) -| node[lab, pos=0.27, above] {Current estimate of the map} (sv.north);
\coordinate (wa) at ($(lm.north)+(-0.8,0)$);
\coordinate (wb) at ($(sv.south)+(0.5,0)$);
\draw[arr] (wa) -- ++(0,0.5) coordinate (wc) -- node[lab, pos=0.55, below] {Withdraw /\\ re-admit} (wc -| wb) -- (wb);
\draw[arr, dashed] ($(sv.south)+(-0.5,0)$) |- node[lab, pos=0.75, above] {Locked on} (lay.east);
\draw[arr] (fe.south) |- (lay.west);
\draw[arr] (lay.north) -- (tm.south);
\end{tikzpicture}
\caption{Overview of the processing flow of BatSLAM 2.0. For every pulse, the binaural echo is converted into a local view $V$ with an energy image $E$ and a spectral-shape image $S$ (the direction cue), which is compared with all stored templates. The five best matches are candidates for the sequence verifier, which keeps several recognition hypotheses alive and commits to one only when it is long, strong, unambiguous and plausible given the current pose uncertainty. A committed hypothesis injects weak loop closure links into the pose graph, which is solved incrementally with iSAM2. The links of every hypothesis form a group that the link management can withdraw from the graph or re-admit later. New templates are anchored to pose-graph nodes, so that their poses follow every optimization, and no templates are laid while the robot is locked onto known territory.}
\label{fig:overview}
\end{figure*}
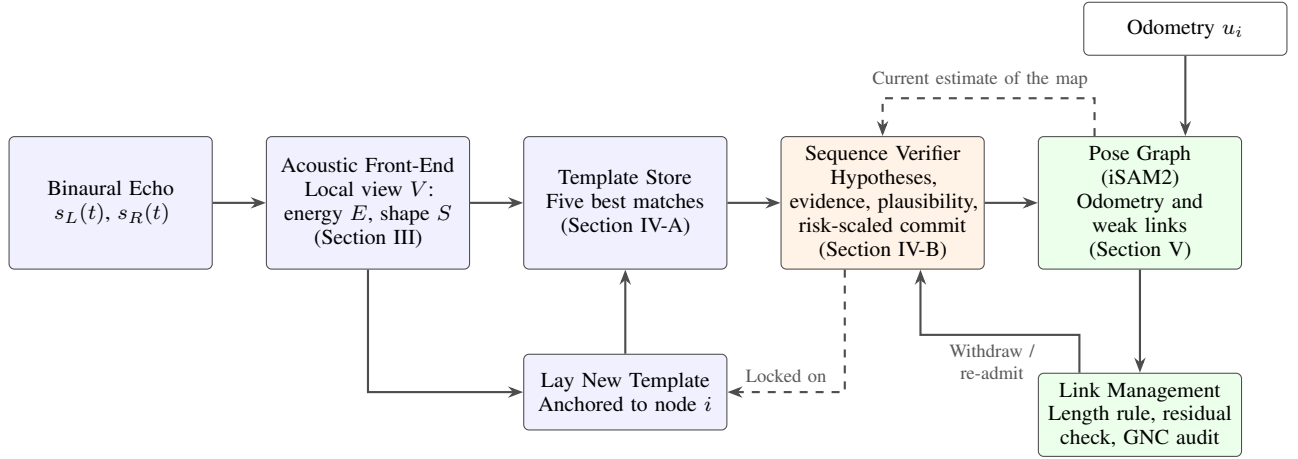

The overall processing flow of BatSLAM 2.0 can be found in figure \ref{fig:overview}, and the accompanying video (footnote \ref{fn:video}) shows it in operation. A broadband signal is emitted by an emitter, and the subsequent echoes are received and processed into a cochlear representation (see \cite{steckelBatSLAMSimultaneousLocalization2013}. The robot's odometry adds a new node to the pose graph for each measurement, connected to the previous node by an odometry factor. Each binaural echo cochleogram is converted into a local view descriptor by the acoustic front-end, which is subsequently compared with all stored templates. The best matches in this comparison become recognition candidates, which are handed to the sequence verifier. The sequence verifier checks whether subsequent measurements give rise to subsequent matches in the pose graph, which is a strong signal for recognizing correct loop closures, and then generates a loop closure hypothesis. When the verifier commits to a hypothesis, its matches become weak loop closure links in the pose graph, grouped per hypothesis, which can be withdrawn in a later stage by the link management subsystem in case of competing evidence becoming available. 

\section{The acoustic front-end}
\label{sec:frontend}
In this section, we describe how a binaural echo train is converted into the local view that is used for place recognition. We start from a model of the received signals, after which we describe the cochlear processing and the two images that make up the local view. The processing steps are illustrated on a simulated pulse in figure \ref{fig:frontend}.

\subsection{Echo formation}
The robot emits a broadband call $s_e(t)$ (in our case a downward frequency-modulated sweep from \SI{90}{\kilo\hertz} to \SI{30}{\kilo\hertz}) and receives the echoes with two ears. During emission, the call is filtered by the directivity of the emitter transducer, modeled as an impulse response $h_e(t,\psi)$ for every direction $\psi$. Every reflector adds its own filtering $h_{r,n}(t)$, and upon reception, the echo is filtered by the directivity of the left and right ear, $h_L(t,\psi)$ and $h_R(t,\psi)$. For $N$ reflectors, the signal received by the left ear can then be written as follows:
\begin{equation}
\begin{aligned}
s_L(t) = \sum_{n=1}^{N} a_n \cdot {}& h_L(t,\psi_n) * h_{r,n}(t) * h_e(t,\psi_n) \\
& * s_e(t - \tau_n) + w_L(t),
\end{aligned}
\label{eq:echo}
\end{equation}
with $*$ denoting the time-domain convolution, $\psi_n$ the direction of the $n$-th reflector, $\tau_n = 2 r_n / c$ its round-trip delay (with $r_n$ its distance and $c = \SI{343}{\meter\per\second}$), $a_n$ the attenuation due to spherical spreading, and $w_L(t)$ additive sensor noise. The right ear follows by replacing $L$ with $R$ in these equations. The reflector filtering $h_{r,n}(t)$ also includes the frequency-dependent atmospheric absorption along the path \cite{iso9613AcousticsAttenuationSound1993}. In the frequency domain, the combined filtering of emitter and ear becomes a product, the Echolocation-Related Transfer Function (ERTF):
\begin{equation}
\begin{aligned}
H^E_L(f,\psi) &= H_e(f,\psi) \cdot H_L(f,\psi), \\
H^E_R(f,\psi) &= H_e(f,\psi) \cdot H_R(f,\psi),
\end{aligned}
\end{equation}
where $H_e$, $H_L$ and $H_R$ are the Fourier transforms of the impulse responses above, i.e., the directivity of the emitter and the Head-Related Transfer Functions (HRTFs) of both ears. 

\subsection{Cochlear processing}
Following the original BatSLAM \cite{steckelBatSLAMSimultaneousLocalization2013} and the spectrogram correlation and transformation receiver \cite{peremansSpectrogramCorrelationTransformation1998}, we model the cochlea as a bank of bandpass filters followed by an envelope detector. Before that, we compress every echo into a short pulse with a matched filter, i.e., a correlation with the emitted call, which improves the range resolution. For frequency channel $k$ of the left and the right ear, this yields:
\begin{equation}
\begin{aligned}
x_{L,k}(t) &= s_L(t) * s_e(-t) * g_k(t), \\
x_{R,k}(t) &= s_R(t) * s_e(-t) * g_k(t),
\end{aligned}
\end{equation}
where $g_k(t)$ is a bandpass filter with a Gaussian frequency response on a logarithmic frequency axis, centered at $f_k$ and with a width of two channel spacings. We use $K = 64$ channels, logarithmically spaced between \SI{30}{\kilo\hertz} and \SI{90}{\kilo\hertz}. The envelope of every channel is the magnitude of its analytic signal, obtained with the Hilbert transform $\mathcal{H}$ \cite{marpleComputingDiscretetimeAnalytic1999}:
\begin{equation}
\begin{aligned}
e_{L,k}(t) &= \sqrt{x_{L,k}(t)^2 + \mathcal{H}\big(x_{L,k}(t)\big)^2}, \\
e_{R,k}(t) &= \sqrt{x_{R,k}(t)^2 + \mathcal{H}\big(x_{R,k}(t)\big)^2}.
\end{aligned}
\end{equation}
In practice, we compute the filtering and the Hilbert transform in one step, by keeping only the positive frequencies of each channel in the frequency domain. As every channel is narrow, its envelope can also be computed from a short inverse Fourier transform of only the frequency bins that the channel occupies, which yields the envelope at a decimated rate and reduces the cost of the front-end by a factor of about 15 (from roughly \SI{300}{\milli\second} to \SI{25}{\milli\second} of compute time per pulse in our Python implementation, at a correlation of 0.999 with the full-rate computation).

Next, the envelopes are converted to range $r = c \cdot t / 2$, and their power is averaged in range bins of \SI{6}{\centi\meter} between \SI{0.25}{\meter} and \SI{10}{\meter} (162 bins), and over pairs of neighboring channels (32 channels). We write the result as $P_L[k,b]$ and $P_R[k,b]$, with $k$ the channel and $b$ the range bin.

\begin{figure*}[t]
\centering
\includegraphics[width=\linewidth]{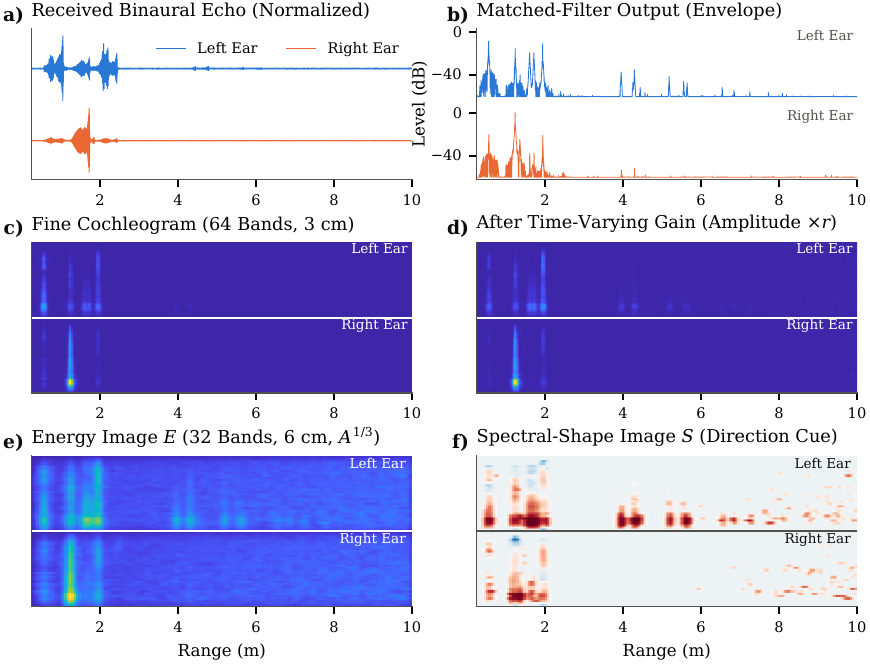}
\caption{Processing steps of the acoustic front-end, illustrated on pulse 2600 of the development drive on the indoor floor. Panel a) shows the received binaural echo train, normalized per ear, as a function of range. Panel b) shows the output of the matched filter for the left ear in decibels, which reveals weak echoes up to \SI{10}{\meter} that are invisible in the raw signal. Panel c) shows the fine cochleogram (64 bands between \SI{30}{\kilo\hertz} and \SI{90}{\kilo\hertz}, \SI{3}{\centi\meter} range bins) as linear amplitude, for both ears (bands low to high within each ear), and panel d) shows the same cochleogram after the time-varying gain. Panel e) shows the energy image $E$ of the descriptor (32 pooled bands, \SI{6}{\centi\meter} range bins, cube-root compression), and panel f) the spectral-shape image $S$ (red: bands above the mean level of the range bin, blue: below), which carries the direction-dependent filtering of the ears and the emitter. The speckle in panel f) beyond \SI{5}{\meter} is noise that passes the level mask.}
\label{fig:frontend}
\end{figure*}

\textbf{Time-varying gain:} Echoes from far away are weaker than echoes from nearby objects by tens of decibels (figure \ref{fig:frontend}, panel b)). Bats compensate for this partially, by reducing their hearing sensitivity right after each call and restoring it over time. We apply a similar time-varying gain (TVG), which multiplies the amplitude of every range bin with its range $r_b$, compensating one of the two spreading factors:
\begin{equation}
\tilde{P}_L[k,b] = r_b^2 \cdot P_L[k,b], \qquad \tilde{P}_R[k,b] = r_b^2 \cdot P_R[k,b].
\end{equation}

\subsection{The local view}
After these initial processing steps, we construct the local view which consists of two images per ear, each of $32 \times 162$ pixels (frequency channel, range bin). The first one, which we call the \emph{energy image}, is the echo amplitude relative to the strongest bin of the view, compressed with a cube root:
\begin{equation}
\begin{aligned}
E_L[k,b] &= \left( \frac{\tilde{P}_L[k,b]}{P_{\max}} \right)^{1/6}, \\
E_R[k,b] &= \left( \frac{\tilde{P}_R[k,b]}{P_{\max}} \right)^{1/6},
\end{aligned}
\end{equation}
where $P_{\max}$ is the maximum of $\tilde{P}$ over both ears, all channels and all range bins, and the power $1/6$ corresponds to the cube root of the amplitude (which as an additional square root term). In contrast to the logarithmic compression of the original BatSLAM, this compression has no floor parameter that needs to be adapted to the noise level of the sensor.

The second image, which we call the \emph{spectral-shape image}, captures the direction cue of the echoes, which is dominated by the spectral content of the received echo. As shown in equation \ref{eq:echo}, every echo is filtered by the ERTF of the direction it arrives from, which gives it a characteristic spectrum. To separate this spectral shape from the level of the echo, we subtract, per range bin, the mean level over all channels:
\begin{equation}
\begin{aligned}
S_L[k,b] &= \frac{D_L[k,b] - \bar{D}_L[b]}{\SI{15}{\deci\bel}}, \\
S_R[k,b] &= \frac{D_R[k,b] - \bar{D}_R[b]}{\SI{15}{\deci\bel}},
\end{aligned}
\label{eq:shape}
\end{equation}
where $D_L[k,b] = 10 \log_{10} \big( \tilde{P}_L[k,b] / P_{\max} \big)$ is the level of a bin in decibels (and likewise $D_R$), and $\bar{D}_L[b]$ and $\bar{D}_R[b]$ are the means over the 32 channels. We clip both images to the interval $[-1, 1]$, and set it to zero in bins that are more than \SI{40}{\deci\bel} below $P_{\max}$, as these contain no echo energy. Both images are smoothed along range with a Gaussian kernel of one bin (\SI{6}{\centi\meter}), so that a template tolerates a few centimeters of displacement.

Finally, the energy images of both ears are concatenated into $E = (E_L | E_R)$, and the spectral-shape images into $S = (S_L | S_R)$, with $|$ denoting concatenation. Each of both is normalized to zero mean and unit energy, and together they form the local view:
\begin{equation}
V = \left( \frac{E - \mu_E}{\| E - \mu_E \|} \; \bigg| \; \frac{S - \mu_S}{\| S - \mu_S \|} \right),
\end{equation}
with $\mu_E$ and $\mu_S$ the means of both images. This gives both images the same weight, independently of their dynamic range. The local view $V$ has 20,736 elements (\SI{83}{\kilo\byte} in single precision).

\begin{figure*}[t]
\centering
\includegraphics[width=\linewidth]{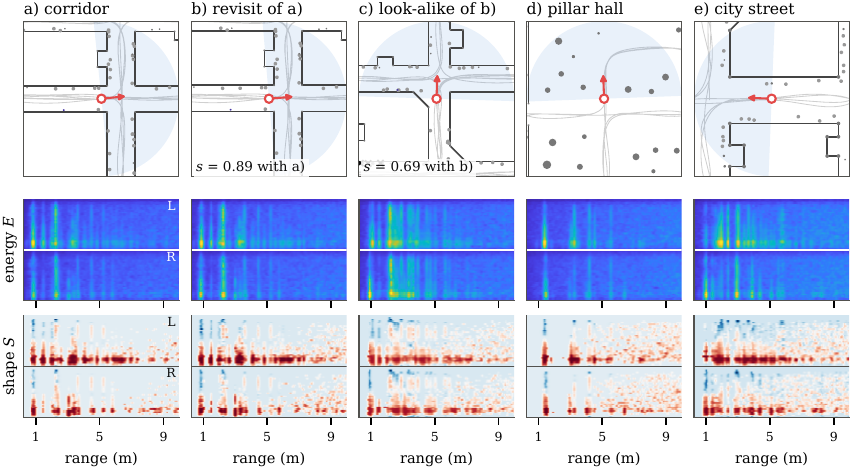}
\caption{Snapshots of five places and their local views. The top row shows the scene within \SI{5}{\meter} of the robot (circle, with an arrow for its heading), with the frontal half-space of the sonar shaded in blue and the complete route in light grey. The middle row shows the energy image $E$ and the bottom row the spectral-shape image $S$ of the local view (red: above the mean level of the range bin, blue: below), each with the left ear above the right ear. Panels a) and b) show the same place in a corridor of the indoor floor, visited \SI{641}{\meter} of travel apart; panel c) shows the view of another place, \SI{7.7}{\meter} away, that is most similar to b) among all earlier pulses of the drive; panel d) shows the pillar hall of the indoor floor, and panel e) a street in the city. The similarity $s$ is 0.89 between a) and b), and 0.69 between b) and c).}
\label{fig:scenes}
\end{figure*}

Figure \ref{fig:scenes} shows the local views of five places, together with the scene around the robot. Panels a) and b) show the same place in a corridor of the indoor floor, on two visits that are \SI{641}{\meter} of travel apart. Both local views are nearly identical, with a similarity of 0.89 (equation \ref{eq:similarity} below). Panel c) shows the view of another place that is most similar to b), among all earlier pulses of the drive. At first sight, it looks alike, as both views are dominated by the echoes of the walls of a junction, but the echoes arrive at different ranges, and their spectral shapes differ, which yields a similarity of only 0.69. Panels d) and e) show a place in the pillar hall, where the local view consists of the isolated echoes of individual pillars, and a place in a street of the city world.

\section{Recognizing places}
\label{sec:recognition}
In this section, we describe how BatSLAM 2.0 recognizes a place that the robot has visited before. We first describe the local view templates in more detail and how they are compared with the stored local view templates. After this description, we introduce the sequence verifier which turns individual, ambiguous matches into verified loop closures.

\subsection{Pose-anchored local view templates}
\label{sec:templates}
A local view template is a local view $V_t$ that is stored together with the index of the pose-graph node at which it was recorded (its \emph{anchor}). The pose of a template is therefore always the current estimate $(x, y, \theta)$ of its anchor node, and it moves with every optimization of the full pose graph. As sonar views are strongly direction dependent (section \ref{sec:resultsCapture}), the heading needs to be part of the template's pose. A new template is laid whenever the robot has traveled \SI{0.45}{\meter} or turned \SI{20}{\degree} since the previous one, to overcome the buildup of templates/poses that are all geometrically located at the same place.

\textbf{Similarity:} We compare the current local view $V_q$ with a template $V_t$ using the correlation coefficient $\rho$, as in our earlier work on RadarSLAM \cite{schoutenBiomimeticRadarSystem2019}, as we found that this was a more robust metric compared to the euclidean distance in the original BatSLAM work. As a revisit never passes through exactly the same position, we shift the range axis of $V_q$ by up to three range bins (\SI{\pm 18}{\centi\meter}) during comparison, and keep the best match:
\begin{equation}
s(q,t) = \max_{|\delta| \le 3} \rho\big( V_q^{(\delta)}, V_t \big),
\label{eq:similarity}
\end{equation}
with $s(q,t)$ the similarity between a query $q$ (the current pulse) and a template $t$, and $V_q^{(\delta)}$ the local view of the query, shifted by $\delta$ range bins$s(q,t)$. The winning shift $\delta^*$ also indicates where the robot is with respect to the anchor, and we use it as a forward offset $d^* = \delta^* \cdot \SI{6}{\centi\meter}$ in the loop closure links.

\textbf{Candidates:} For every pulse, the five most similar templates become candidates, provided that $s(q,t) \ge \Smin$, that the template was laid more than \SI{6}{\meter} of travel earlier, and that the headings of the robot and the anchor differ by less than \SI{69}{\degree}. No position gate is applied, i.e., the search over the templates is global.

\subsection{The sequence verifier}
\label{sec:verifier}
A single match in the sonar data is never trusted, as sonar data can be highly ambiguous. Instead, we keep a set of \emph{recognition hypotheses}, and each of these hypotheses claims that the stretch of path that is being driven now was driven before. A hypothesis $h$ is a chain of $n$ pairs $(q_i, t_i)$ of query nodes and templates, in which every pair states that the anchor of $t_i$ lies $d^*$ in front of $q_i$, with the same heading.

\textbf{Geometric consistency:} A new pair can only extend a hypothesis if the motion between the queries agrees with the motion between the anchors. For two pairs $a$ and $b$, let $\Delta p_q$ and $\Delta\theta_q$ be the displacement and heading change between both queries according to odometry, and $\Delta p_t$ and $\Delta\theta_t$ the same quantities between both anchors according to the map. Pair $b$ is consistent with pair $a$ when:
\begin{equation}
\begin{aligned}
\| \Delta p_t - \Delta p_q \| &< \SI{0.30}{\meter} + 0.15 \cdot \| \Delta p_q \|, \\
|\Delta\theta_t - \Delta\theta_q| &< \SI{0.25}{\radian},
\end{aligned}
\label{eq:consistency}
\end{equation}
in which the position tolerance grows with the distance traveled, as the position error of odometry is unbounded and grows with every meter driven, both directly and through the accumulated heading error. The heading error, in contrast, stays well below \SI{0.25}{\radian} over the length of a hypothesis (a few meters), so that a fixed heading tolerance suffices. We test every new pair against both the last and the first pair of the hypothesis. Every pulse, each hypothesis is extended with its best consistent candidate, and every unused candidate seeds a new hypothesis. Several contradicting hypotheses can therefore be alive at the same time, and this set of hypotheses represents the ambiguity.

\textbf{Evidence:} Every hypothesis accumulates evidence as a sum of log-likelihood ratios of its similarities, minus a penalty for every pulse without a consistent candidate:
\begin{equation}
\Lambda(h) = \sum_{i=1}^{n} \ell\big(s(q_i, t_i)\big) - 0.5 \cdot m,
\label{eq:evidence}
\end{equation}
with $m$ the number of missed pulses. The term $\ell(s)$ is the evidence that a single pair with similarity $s$ provides for a true revisit. Ideally, it is the log-likelihood ratio, which by Bayes' rule equals the posterior minus the prior log-odds of a correct candidate:
\begin{equation}
\begin{aligned}
\ell(s) &= \ln \frac{p(s \mid \text{correct})}{p(s \mid \text{wrong})} \\
&= \operatorname{logit} P(\text{correct} \mid s) - \operatorname{logit} P(\text{correct}),
\end{aligned}
\label{eq:llr}
\end{equation}
with $p(s \mid \text{correct})$ and $p(s \mid \text{wrong})$ the distributions of the similarity of candidates whose anchor does and does not lie at the current place of the robot (within \SI{0.6}{\meter} and \SI{26}{\degree} according to ground truth), and $P(\text{correct})$ the fraction of correct candidates. A positive $\ell$ therefore favors a revisit, and a negative $\ell$ favors a different place. We fitted $P(\text{correct} \mid s)$ once, with a logistic regression $\operatorname{logit} P(\text{correct} \mid s) = a \cdot s + b$ on the labeled candidates of the development drive. The log-likelihood ratio is then linear in $s$, and we refer to it, clipped, as the \emph{evidence curve}:
\begin{equation}
\ell(s) = \min\Big(4, \max\big(-4,\ \LLRslope \cdot (s - \LLRmid)\big)\Big).
\label{eq:evidenceCurve}
\end{equation}
Here, the slope \LLRslope{} is the fitted $a$, i.e., the evidence gained per unit of similarity, and \LLRmid{} $= (\operatorname{logit} P(\text{correct}) - b)/a$ is the similarity at which a candidate is as likely to be correct as any other candidate, so that it carries no evidence ($\ell = 0$). The clipping to $[-4, 4]$, reached at $s = \LLRmid \pm 4/\LLRslope$, prevents a single exceptionally good or bad match from dominating a hypothesis, as the linear model extrapolates without bound into the tails, where few candidates exist. A hypothesis dies after four consecutive missed pulses.

\textbf{Commit rule:} A hypothesis is committed (i.e., turned into loop closure links) when it has at least 8 pairs on at least 3 templates, an evidence $\Lambda \ge 10$, and a margin of at least 4 over the best rival hypothesis that places the robot elsewhere (more than \SI{1}{\meter} or \SI{0.4}{\radian} away). Furthermore, the correction it implies for the map must be plausible, which is estimated by a plausibility gate.

\textbf{Plausibility gate:} A hypothesis can be consistent and still be wrong, e.g., when a corridor has the same structure as another corridor a few meters away. Committing such an alias would force a jump of the current pose that is much larger than the drift the odometry could have accumulated. We therefore test whether the correction implied by a hypothesis can be explained by the uncertainty of the current pose estimate. From its last pair, the hypothesis predicts the current pose $x_h$, which we compare with the current estimate $\hat{x}$ of the pose graph through their Mahalanobis distance:
\begin{equation}
d(h) = (x_h - \hat{x})^T \cdot \Sigma^{-1} \cdot (x_h - \hat{x}) \le 16.27,
\label{eq:plausibility}
\end{equation}
where $\Sigma$ describes how far $x_h$ and $\hat{x}$ may disagree when the hypothesis is correct. Two independent errors contribute: the drift of the estimate $\hat{x}$, given by its covariance in the pose graph, and the error of $x_h$, which stems from a single sonar match and therefore has the covariance of one loop closure link. As both errors are independent, their covariances add. For a correct hypothesis, $d(h)$ then follows a $\chi^2$ distribution with three degrees of freedom ($x$, $y$, $\theta$), and 16.27 is its 99.9th percentile: a larger $d(h)$ means that the jump cannot be explained by drift and match error, and the hypothesis is rejected. For small corrections, we use the marginal covariance of the current node, which iSAM2 provides cheaply. This marginal, however, includes the drift that the current node shares with the anchor. For corrections larger than \SI{1.5}{\meter}, we therefore use the covariance of the relative pose between the anchor and the current node, obtained from their joint marginal. Once loops have been closed, this relative covariance is much smaller, and corrections of several meters become implausible.

\textbf{Risk-scaled commit:} The damage of a wrong commit grows with the correction it forces on the map. When the implied correction exceeds \SI{1.5}{\meter}, we therefore require at least 16 pairs (about \SI{2.4}{\meter} of travel), an evidence $\Lambda \ge 40$ and a mean evidence of at least one per pair ($\Lambda \ge n$). Brief aliases die before they reach this length and strength, and the last condition stops the rare persistent alias along quasi-periodic structures (such as a lane between two rows of pillars), which stays consistent for many meters but matches only weakly (section \ref{sec:resultsAblation}).

\textbf{Lock-on:} Once a hypothesis is committed, all its pairs are released at once as loop closure links, which enter the pose graph as one link group (section \ref{sec:backend}). Rival hypotheses that agree with it are discarded, as they describe the same revisit. The committed hypothesis itself stays alive, and the verifier keeps extending it with the same consistency test (equation \ref{eq:consistency}). Every new pair is released immediately as an additional link of the same group, without passing the commit rule again: the pair continues a chain that has already been verified, and its consistency with both the first and the last pair ties it to that chain. At every pulse at which a committed hypothesis is extended, we say that the robot is \emph{locked on} to a known stretch of path. Lock-on thus produces a dense sequence of links along the whole revisit, rather than a single burst at the moment of commitment, so that the drift is corrected along the entire revisited stretch. While locked on, no new templates are laid, as the robot drives through a place that is already represented by templates, and new templates of the same place would only compete with the existing ones as candidates at later revisits. Lock-on ends when the committed hypothesis dies, i.e., when the robot leaves the known path and no longer finds consistent candidates, or when the link management of the back-end withdraws its link group (section \ref{sec:backend}). From then on, templates are laid again. The complete processing of a pulse is summarized in algorithm \ref{alg:pulse}.

\begin{algorithm}[t]
\caption{Processing of one pulse in BatSLAM 2.0.}
\label{alg:pulse}
\small
\begin{algorithmic}[1]
\State Add a node with odometry to the pose graph
\State Compute the local view from the echo \Comment{Section \ref{sec:frontend}}
\State Update the recognition hypotheses \Comment{Section \ref{sec:verifier}}
\State Commit a strong, unambiguous and plausible hypothesis
\State Release its links and manage link groups \Comment{Section \ref{sec:backend}}
\State Lay a new template, unless locked on
\end{algorithmic}
\end{algorithm}

\section{The robust pose graph}
\label{sec:backend}
In this section, we describe how odometry and verified recognitions are fused into a one consistent topological map. In contrast to the pose cells and experience map of the original BatSLAM, a pose graph gives a principled estimate of every pose and its uncertainty, and it allows loop closures to be added and removed at any time, as was described in the previous section. In this section, we will detail the exact implementation of the pose graph subsystem with the GTSAM backend framework.

\subsection{Formulation}
Every pulse $i$ adds a node with the pose $x_i = (x, y, \theta)$ of the robot, connected to the previous node by the odometry increment $u_i$. A committed recognition adds a link between a query node $q$ and the anchor $a$ of the matched template, with the measurement $z_{qa} = (d^*, 0, 0)$. The map is then the solution of:
\begin{equation}
\begin{aligned}
\hat{X} = \arg\min_{X} \; & \sum_{i} \big\| \Delta(x_{i-1}, x_i) - u_i \big\|^2_{\Sigma_{o,i}} \\
+ & \sum_{(q,a) \in \mathcal{L}} \rho_H\Big( \big\| \Delta(x_q, x_a) - z_{qa} \big\|_{\Sigma_\ell} \Big),
\end{aligned}
\label{eq:map}
\end{equation}
where $\Delta(x_a, x_b)$ is the pose of node $b$ in the frame of node $a$, $\|\cdot\|_\Sigma$ is the Mahalanobis norm, $\mathcal{L}$ is the set of loop closure links currently in the graph, and $\rho_H$ is the Huber loss. The odometry uncertainty $\Sigma_{o,i}$ grows with the square root of the distance travelled (appendix \ref{app:params}). The links are deliberately weak ($\sigma = \SI{0.30}{\meter}$ in position and \SI{0.20}{\radian} in heading), so that a single link can only nudge the map, whereas only a verified sequence of links (originating from the sequence verifier) closes a loop. We solve equation \ref{eq:map} incrementally with iSAM2 \cite{kaessISAM2IncrementalSmoothing2012} in GTSAM \cite{dellaertFactorGraphsGTSAM2012}, using Dogleg steps, as Gauss-Newton steps diverged in our system when many conflicting links were present.

\subsection{Link management}
To maximally build safeguards into the system against map collapses, we assume that even a well-verified recognition can still be wrong. Therefore, we included a system to allow the pose graph to be able to forget these erroneous links. We therefore keep the links of every committed hypothesis together as a \emph{link group}, which enters the graph as tentative, and can be removed from iSAM2 and re-admitted later. Three rules act on these groups:
\begin{itemize}
\item \textbf{Length rule:} a group whose hypothesis ends with fewer than 16 links is withdrawn permanently. This rule targets the same brief aliases as the risk-scaled commit, but for hypotheses with a small implied correction.
\item \textbf{Residual check:} after every injection of links, the tentative group with the worst median Mahalanobis residual is removed if that median exceeds 12.
\item \textbf{GNC audit:} every 400 nodes, we solve equation \ref{eq:map} in batch with Graduated Non-Convexity (GNC) \cite{yangGraduatedNonConvexityRobust2020}, over the odometry and the links of \emph{all} groups, including withdrawn ones. A group is kept (or re-admitted) when the mean GNC weight of its links is at least 0.5, and a group that survives two audits is confirmed.
\end{itemize}
The first two rules judge every group by itself, whereas the GNC audit operates globally on the graph, judging all groups jointly, similar to the RRR method \cite{latifRobustLoopClosing2013} or the switchable constraints \cite{sunderhaufSwitchableConstraintsRobust2012}, but per group rather than per link. As we will show in section \ref{sec:resultsAblation}, the audit actually never changed a final map in our experiments, making it largely reduntant at this point. However, we opted to keep this mechanism present as a safeguard for more strongly aliased environments.

\section{Simulation setup}
\label{sec:setup}
In this section, we describe the approach used to simulate the sensor system, the worlds, the odometry model and the metrics used to evaluate BatSLAM 2.0.

\subsection{Sonar simulator}
\begin{figure*}[t]
\centering
\includegraphics[width=0.86\linewidth]{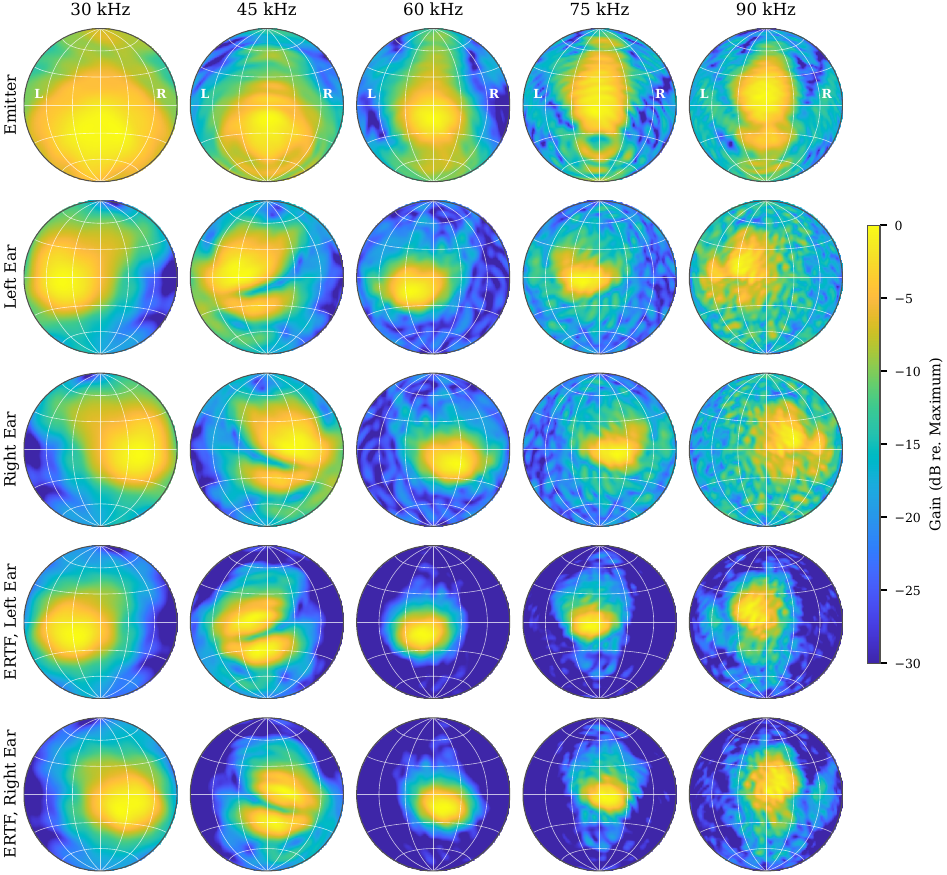}
\caption{Directivity of the simulated sonar (Lambert azimuthal equal-area projection of the frontal hemisphere, seen from behind the head; grid lines every \SI{30}{\degree}). Rows: emitter, left ear, right ear and the ERTFs of both ears, at five frequencies between \SI{30}{\kilo\hertz} and \SI{90}{\kilo\hertz}, each normalized to its maximum.}
\label{fig:hrtf}
\end{figure*}

We simulated the binaural sonar by implementing equation \ref{eq:echo} directly. The emitted call is a \SI{3}{\milli\second} linear FM downsweep from \SI{90}{\kilo\hertz} to \SI{30}{\kilo\hertz}, sampled at \SI{500}{\kilo\hertz}. For every reflector and ear, the simulator combines the directivity of the emitter and the ear (taken from the simulated HRTF of \textit{Phyllostomus discolor} \cite{demeySimulatedHeadRelated2008}, shown in figure \ref{fig:hrtf}), spherical spreading, atmospheric absorption \cite{iso9613AcousticsAttenuationSound1993} and the reflectivity of the reflector. Walls and cylinders reflect specularly, and point reflectors isotropically. White Gaussian noise is added to both ears, and we express its standard deviation relative to the peak amplitude of the emitted call. Unless stated otherwise, we use the design sensor, with a noise level of \SI{-116}{\deci\bel} and a listening window of \SI{10}{\meter} (section \ref{sec:resultsFrontend}). 

\subsection{Worlds and drives}
\begin{figure*}[t]
\centering
\includegraphics[width=\linewidth]{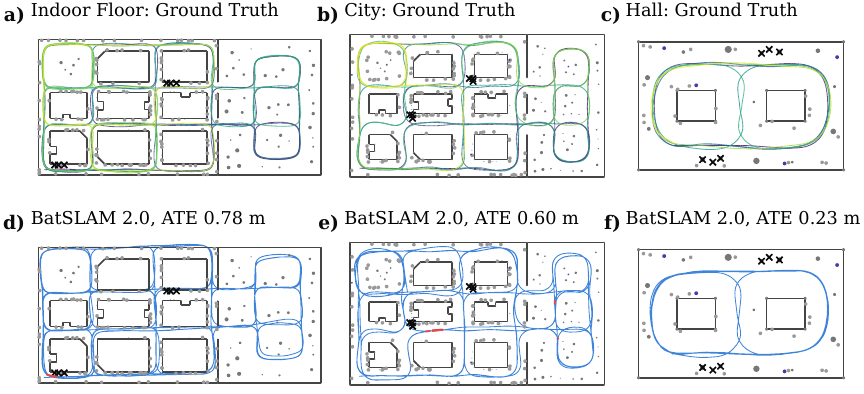}
\caption{The simulated worlds and exemplary results of BatSLAM 2.0. Top row: the ground-truth drive (colored from start to end) in a) the indoor floor (\SI{38}{\meter} by \SI{18}{\meter}, development drive on route 3, \SI{964}{\meter}), b) the city (\SI{46}{\meter} by \SI{26}{\meter}) and c) the small development hall (\SI{16}{\meter} by \SI{10}{\meter}). Black crosses: the two copies of the aliasing pattern. Bottom row, d)--f): the final map estimated by BatSLAM 2.0 for the same drives with calibrated odometry (blue), with its trajectory error (ATE); wrong links in the final graph are drawn in red.}
\label{fig:worlds}
\end{figure*}

We used three simulated worlds (figure \ref{fig:worlds}). The main world is an \emph{indoor floor} of \SI{38}{\meter} by \SI{18}{\meter}, consisting of a grid of rooms separated by corridors of \SI{1.4}{\meter} to \SI{1.8}{\meter} wide, and a hall with 40 pillars. The walls carry 170 small reflectors (e.g., pipes and radiators), and to make aliasing explicit, we placed an identical pattern of three objects on two different corridor walls. The second world is a \emph{city} of \SI{46}{\meter} by \SI{26}{\meter} with the same topology but with streets of \SI{3}{\meter} to \SI{4}{\meter}, and the third is a small \emph{hall} used during development.

In the indoor floor and the city, the robot drives a random tour of about \SI{1}{\kilo\meter}, in which every corridor is driven at least twice in both directions, with a small lateral along the corridor wobble so that revisits are never exactly identical in position. A pulse is emitted every \SI{0.15}{\meter} of travel, yielding about 6400 pulses per drive. We developed and tuned the system on one route of the indoor floor, and used other random routes as unseen test cases. 

\subsection{Odometry model}
The odometry was generated from the ground-truth increments with random noise (\SI{3}{\percent} forward, \SI{1}{\percent} lateral and \SI{1}{\degree} per $\sqrt{\si{\meter}}$ in heading) and a systematic error (a heading bias of \SI{0.6}{\degree\per\meter} and a scale error of \SI{2}{\percent}). We scaled the systematic part with a bias factor $b$: $b = 1$ represents uncalibrated wheel odometry, $b = 0.25$ a roughly calibrated robot and $b = 0$ a calibrated robot. Over \SI{1}{\kilo\meter}, dead reckoning drifts by \OdoDriftRange{}, depending on $b$ and the random seed. Typical robots in indoor environments can be assumed to have decently calibrated odometry, whereas in outdoor environments these calibrations are less accurate due to a higher prevalence of wheel slippage.

\subsection{Metrics}
As BatSLAM 2.0 builds a topological map, our primary criterion is the absence of wrong loop closures and map collapse rather than metric accuracy. The \emph{link precision} is the fraction of links for which query and anchor are truly within \SI{0.6}{\meter} and \SI{26}{\degree} of each other. The number of \emph{wrong commits} counts committed hypotheses of which the majority of links are wrong. The \emph{revisit coverage} is the fraction of true same-direction revisits that received a correct link. The \emph{collapsed pairs} are the fraction of node pairs that are at least \SI{4}{\meter} apart in reality, but less than half that distance apart in the map. Finally, we report the \emph{absolute trajectory error} (ATE) without alignment and after a rigid alignment in the plane.

The pose-graph part of BatSLAM 2.0 is implemented in Python using GTSAM 4.2 with Python bindings, and the simulator is implemented using custom MATLAB code. All experiments ran on a desktop computer with an AMD Ryzen 9 7900X.

\section{Results}
\label{sec:results}
In this section, we present the results in the order of the processing chain. We first analyze the recognition of a single sonar view, and how far a robot may deviate from a template for it still to be recognized. Next, we evaluate the complete system, followed by ablations, a robustness analysis and the computational cost.

\subsection{Sonar Template Recognition}
\label{sec:resultsFrontend}
To evaluate the front-end on its own, we tested how well a single local view recognizes the place where it was taken, without the sequence verifier. For every third pulse, we compared its local view with those of all earlier pulses at least \SI{6}{\meter} of travel back, using the similarity of equation \ref{eq:similarity}. Only pulses with a true revisit, i.e., an earlier pulse within \SI{0.6}{\meter} and \SI{26}{\degree}, serve as queries, so that only same-direction revisits are counted. We report two metrics. The \emph{top-1 accuracy} is the fraction of queries for which the most similar earlier view lies within \SI{1}{\meter} and \SI{30}{\degree} of the query. The \emph{AUC} is the area under the ROC curve that separates same-place pairs (within \SI{0.6}{\meter} and \SI{26}{\degree}) from different-place pairs (more than \SI{3}{\meter} or \SI{57}{\degree} apart), i.e., the probability that a random same-place pair is more similar than a random different-place pair. We estimate it from 20\,000 random pairs of each class, so that differences in AUC below about 0.005 lie within its sampling noise. Figure \ref{fig:sensor} reports both metrics for different sensors, and table \ref{tab:descriptor} in the appendix for the design choices of the descriptor.

\textbf{Sensor quality:} Figure \ref{fig:sensor} compares four sensors on the same drive through the city map, combining two maximal ranges (\SI{5}{\meter} and \SI{10}{\meter}) with two noise levels (\SI{-96}{\deci\bel} and \SI{-116}{\deci\bel}, i.e., the standard deviation of the noise relative to the peak amplitude of the emitted call). For every sensor, we evaluated three descriptors: the descriptor of the original BatSLAM (8 frequency channels, \SI{3}{\centi\meter} range bins and a logarithmic compression with a floor at \SI{-30}{\deci\bel}), the same with the time-varying gain (TVG), and our final descriptor (section \ref{sec:frontend}). The noise level clearly dominates. At a noise level of \SI{-96}{\deci\bel}, no descriptor reaches a top-1 accuracy of 0.5, and a longer range does not help, as the echoes beyond \SI{5}{\meter} are buried in noise. The TVG then even hurts, as it amplifies this noise, and the final descriptor offers no clear advantage over the original one. With \SI{20}{\deci\bel} less noise (\SI{-116}{\deci\bel}), every descriptor improves substantially, and both the TVG and the final descriptor pay off. Only then does a longer range pay off as well, up to a top-1 accuracy of \SensorTopOneBest{} for the final descriptor at \SI{10}{\meter}. This sensor (\SI{10}{\meter}, \SI{-116}{\deci\bel}) is the design sensor that we use in the remainder of this paper. To test the robustness, we also use three degraded sensors: the design sensor with \SI{10}{\deci\bel} and \SI{20}{\deci\bel} more noise (the \SI{-10}{\deci\bel} and \SI{-20}{\deci\bel} sensors, at \SI{-106}{\deci\bel} and \SI{-96}{\deci\bel}), and the sensor of the original configuration (\SI{5}{\meter}, \SI{-96}{\deci\bel}).

\begin{figure*}[t]
\centering
\includegraphics[width=\linewidth]{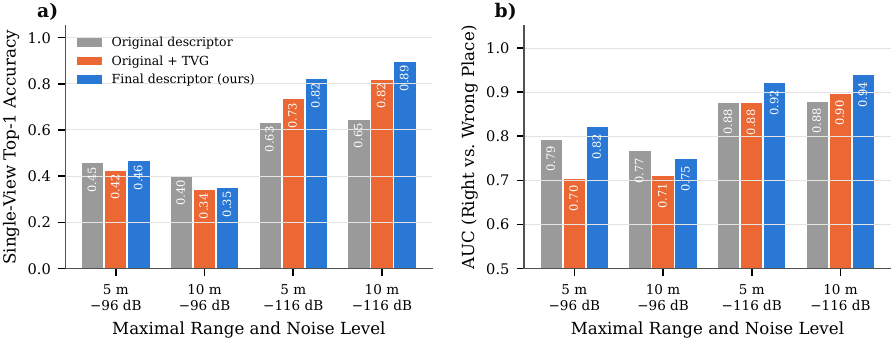}
\caption{Single-view place recognition for four sensors on the city drive: a) top-1 accuracy, b) AUC. Grey: descriptor of the original BatSLAM; orange: the same with TVG; blue: the final descriptor.}
\label{fig:sensor}
\end{figure*}

\textbf{Descriptor design:} Table \ref{tab:descriptor} in the appendix compares the design choices of the descriptor for the design sensor on both worlds. All variants share the spatiospectral resolution (32 frequency channels, \SI{6}{\centi\meter} range bins) and the TVG of our front-end, and differ only in the listed aspect. Adding the spectral-shape image to the energy image raises the top-1 accuracy on the indoor floor from \TopOneEnergyIndoor{} to \TopOneFinalIndoor{}, and the AUC on both worlds. Adding the interaural level difference (ILD) as a third image raises the AUC further, but lowers the top-1 accuracy on both worlds, as the ILD changes quickly with small pose changes, and we therefore left it out. The compression law matters little: with the spectral-shape image, every compressive law reaches a top-1 accuracy between 0.88 and 0.93 on the indoor floor and of 0.89 in the city, and only a linear amplitude is clearly worse. We use the cube root, as it performs on par with a logarithm without a floor that depends on the noise level (section \ref{sec:frontend}). The matching method, in contrast, has shown to matter a lot. With the matching of the original BatSLAM (the Euclidean distance between unit-energy views, without range shift), the top-1 accuracy of our descriptor drops to about 0.80 on both worlds, and that of a logarithmic energy image alone to 0.40 and 0.60. Both the new descriptor and the new matching thus contribute to the improvement over the original BatSLAM. Note that these last rows use the resolution and TVG of our front-end, and therefore differ from the original descriptor in figure \ref{fig:sensor}.

\subsection{Capture region of a view template}
\label{sec:resultsCapture}
\begin{figure}[t]
\centering
\includegraphics[width=\linewidth]{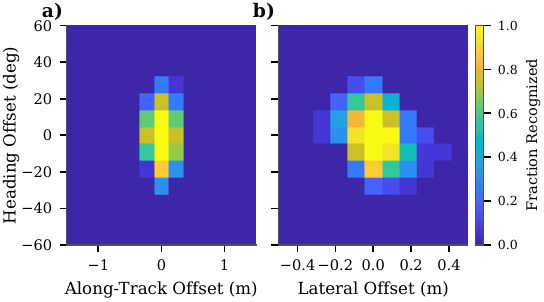}
\caption{Capture region of local view templates on the indoor floor, averaged over 16 template poses: fraction of recognized offset views for combined a) along-track and heading offsets, and b) lateral and heading offsets.}
\label{fig:capture}
\end{figure}

To measure how far a revisit may deviate from a template, we rendered views at controlled offsets from 16 template poses along the development drive. An offset view counts as recognized when its similarity to the template exceeds \LLRmid{} and no template of another place is more similar. Figure \ref{fig:capture} shows that the capture region is narrow: about \CaptureAlong{} along track, \CaptureLateral{} laterally and \CaptureHeading{} in heading. 

\subsection{The complete system}
\label{sec:resultsMain}
\begin{figure}[t]
\centering
\includegraphics[width=0.92\linewidth]{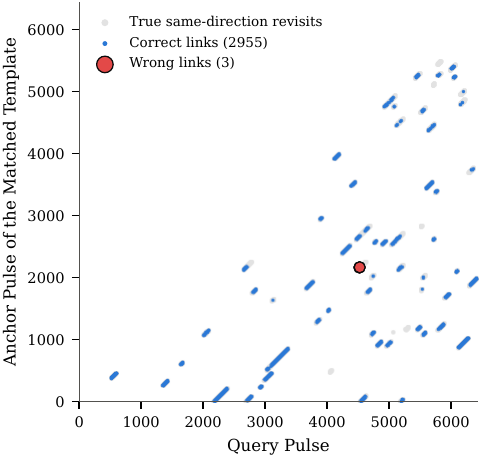}
\caption{Loop closure matrix of the development drive (route 3, calibrated odometry). Every dot is a link in the final graph between a query pulse and the anchor of the matched template. Grey: true same-direction revisits; blue: correct links; red: wrong links.}
\label{fig:loopmatrix}
\end{figure}

We ran the complete system on three odometry seeds of the development drive, the two unseen routes and the city environment (which was unseen during development), each with calibrated odometry and with a residual heading bias, on three degraded sensors, and with uncalibrated odometry. Table \ref{tab:main} in the appendix summarizes the results per condition, figure \ref{fig:worlds}d--f shows the maps of one run per world, figure \ref{fig:loopmatrix} shows the links of one run, and table \ref{tab:mainRuns} and figure \ref{fig:appendixMaps} in the appendix list all runs.

With the design sensor and calibrated or roughly calibrated odometry, all \NdesRuns{} runs produced a consistent map: none of the \CommitsDes{} committed hypotheses was wrong, and no distant places were pulled together. The link precision was \PrecDesRange{}, and \CovDesRange{} of the true revisits were linked. The trajectory error was \ATEdesRange{} (\ATEalignedDesRange{} after alignment), against \ATEodoDesRange{} for odometry alone.

The degraded sensors also produced consistent maps (\ATEalignedDegRange{} after alignment), but on the \SI{-20}{\deci\bel} sensor, one wrong hypothesis of 17 links survived. It was committed early in the drive, when the map was still so uncertain that a correction of \SI{8}{\meter} was plausible, and it distorts the map locally without collapsing it.

The difference between the aligned and unaligned errors is caused by the heading drift that accumulates before the first loop closure, which can only be removed by a later loop closure with that first stretch of the route. If there is none, the whole map remains rotated (by about \SI{5}{\degree} on route 7). The unaligned error therefore measures how the map is anchored, and the aligned error measures its shape. With uncalibrated odometry, the links remain almost all correct, but the metric accuracy degrades (section \ref{sec:resultsRobust}).

\subsection{Ablations}
\label{sec:resultsAblation}
To assess the contribution of every component, we removed or replaced several components of the full system, one at a time, and evaluated all variants on four standard cases (tables \ref{tab:ablation} and \ref{tab:ablationCases} in the appendix). Without sequence verification, the system made 215 wrong commits, of which 10 survived the link management, and the mean trajectory error grew to \SI{4.85}{\meter}. Without link management as well (the naive variant), the map collapsed completely (rendering ATE a useless metric). This confirms that a single sonar view is too ambiguous to be trusted to perform loop closure. The front-end matters has an equally large contribution to the performance of the system, as removing the time-varying gain or the spectral-shape image, wrong hypotheses survive often. Without the range-shift search, the coverage of recognition drops to 0.60. The original front-end and the hand-set evidence constants of our first implementation made no wrong commits, but linked fewer revisits (0.73 and 0.68).

The safeguards against aliases, in contrast, hardly come into play on these cases, as they contain almost no aliases that survive the sequence verification. Removing the complete link management even lowered the error (\SI{0.55}{\meter} against \SI{0.87}{\meter}), as the length rule also withdraws short correct groups. Requiring 24 pairs for a large correction, without a condition on the evidence, raised the error on one particular run to \SI{2.98}{\meter}, as the true revisit of the start of the route was split into two hypotheses that never reached 24 pairs, which left the map rotated.

\textbf{Safeguards against aliases:} To study the safeguards, we repeated the ablations on the four hardest cases: the three degraded sensors and a development drive with a recurring alias (table \ref{tab:ablationHard} in the appendix). Without the risk-scaled commit, 12 of 346 commits were wrong. Of these, 8 implied a correction of more than \SI{1.5}{\meter} (against only \SI{13}{\percent} of the correct commits), and 11 ended with 16 links or fewer, as an alias ends where the similarity between both corridors does. These observations motivate the risk-scaled commit and the length rule. With all safeguards in place, 6 wrong hypotheses were committed, and the link management withdrew all but the early alias on the \SI{-20}{\deci\bel} sensor. Without link management, 7 survived, and removing only the length rule gave identical results: the residual check and GNC audits never changed a final map, as a single wrong hypothesis that passed the gates of the verifier cannot be detected by a joint consistency test. Without any plausibility gate, 3 wrong hypotheses survived. Requiring 24 pairs regardless of the evidence is the only variant without a surviving wrong hypothesis, but as shown above, it also withholds the loop closures that anchor the map (section \ref{sec:limitations}).

\subsection{Robustness}
\label{sec:resultsRobust}
Next, we varied the systematic odometry error with the bias factor $b$, tripled the random odometry noise, and combined the degraded sensors with a residual bias (figure \ref{fig:robustness}; table \ref{tab:robustness} in the appendix). Up to $b = 0.5$ (\SI{0.3}{\degree\per\meter}), every run produced a consistent map with the same coverage as with calibrated odometry, while the error of dead reckoning grew beyond \SI{20}{\meter}. The same holds for three times the random noise (\SI{0.48}{\meter} after alignment). With uncalibrated odometry ($b = 1$), the heading has drifted by about \SI{45}{\degree} at the first revisit. In three of six runs, later loop closures removed this rotation (\SI{1.4}{\meter} to \SI{1.8}{\meter}). In the other three, the links were still correct, but as the pose graph treats odometry errors as random, the stretches between loop closures keep the curvature of the biased odometry, and the map remains distorted (\SI{9.3}{\meter} to \SI{14.5}{\meter}). With $b = 2$, fewer true revisits pass the consistency test and plausibility gate (coverage 0.32 to 0.41), and the map collapses about as much as dead reckoning does, but not a single wrong hypothesis remained. In summary, BatSLAM 2.0 needs odometry calibrated to within about \SI{0.3}{\degree\per\meter}, and beyond that, it degrades in metric accuracy rather than in topology.

\begin{figure*}[t]
\centering
\includegraphics[width=\linewidth]{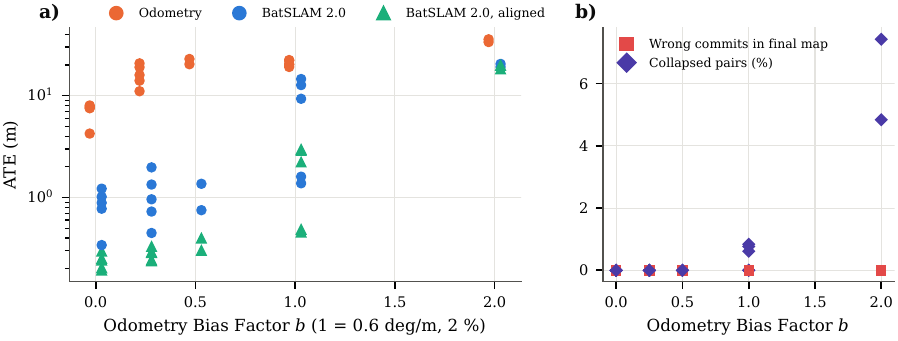}
\caption{Robustness against systematic odometry errors on the development drive. a) Trajectory error of odometry (orange), BatSLAM 2.0 (blue) and BatSLAM 2.0 after alignment (green) versus the bias factor $b$; b) wrong hypotheses in the final graph and percentage of collapsed pairs.}
\label{fig:robustness}
\end{figure*}

\subsection{Computational cost}
\label{sec:resultsCost}
On a single core, a pulse took \SI{\CostTotal}{\milli\second} on average, of which \SI{\CostGraph}{\milli\second} went to the pose graph. As the template comparison and the pose graph grow with the map, our (unoptimized) Python implementation meets the \SI{100}{\milli\second} budget of a \SI{10}{\hertz} pulse rate on average, but not towards the end of a \SI{1}{\kilo\meter} drive.

\section{Real-world experiment}
\label{sec:real}
\begin{figure*}[t]
\centering
\includegraphics[width=\linewidth]{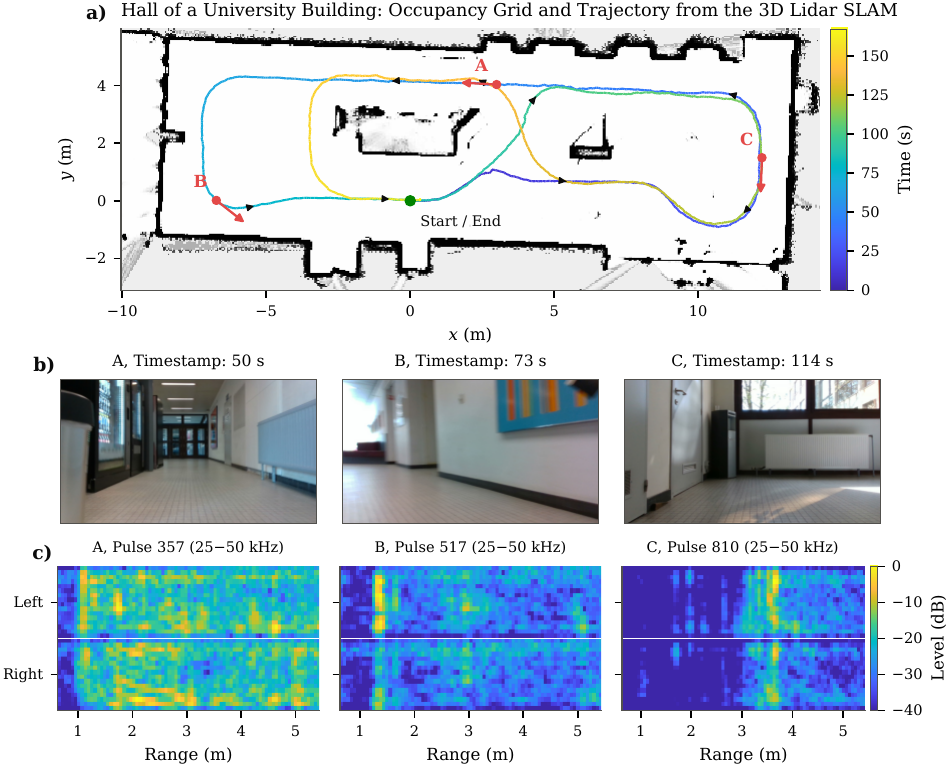}
\caption{The real-world recording. a) Occupancy grid of the hall from the lidar scans, with the ground-truth trajectory (colored by time); the left loop was driven twice in the same direction, the right loop once in each direction. b) Camera images at places A, B and C. c) Binaural cochleograms of the closest sonar pulse, formed with the virtual ears of equation \ref{eq:ears}.}
\label{fig:realWorld}
\end{figure*}

In this section, we apply BatSLAM 2.0 to a real recording. As no binaural bat-head sonar recording of a suitable drive was available, we used an embedded 3D sonar sensor (eRTIS \cite{kerstensERTISFullyEmbedded2019}), and formed two virtual ears from its microphone array. Apart from the virtual ears and the frequency band, the system is identical to the one of the previous sections.

\subsection{Recording, virtual ears and ground truth}
\label{sec:realSetup}
\textbf{Recording:} The eRTIS sensor was mounted on a mobile platform together with a 3D lidar (Ouster OS0) and a camera, and driven through a hall of \SI{22}{\meter} by \SI{8}{\meter} in a university building (figure \ref{fig:realWorld}). The drive is a figure-eight of \RealDist{}~\si{\meter}, during which the sensor emitted \RealPulses{} calls at about \RealRateHz{}~\si{\hertz}. The call is a \SI{2.5}{\milli\second} FM downsweep from \SI{50}{\kilo\hertz} to \SI{25}{\kilo\hertz}, recorded by 32 MEMS microphones in a pseudo-random planar array of about \SI{7}{\centi\meter} by \SI{4}{\centi\meter}. We used the ranges between \SI{0.6}{\meter} and \SI{5.5}{\meter}.

\textbf{Virtual ears:} We split the array into a left and a right half of 16 microphones each, and steered each half with a delay-and-sum beamformer towards an azimuth of \SI{30}{\degree} to its own side:
\begin{equation}
\begin{aligned}
s_L(t) &= \sum_{i \in \mathcal{L}} w_i \, x_i\!\left(t + \frac{y_i \sin 30^\circ}{c}\right), \\
s_R(t) &= \sum_{i \in \mathcal{R}} w_i \, x_i\!\left(t - \frac{y_i \sin 30^\circ}{c}\right),
\end{aligned}
\label{eq:ears}
\end{equation}
with $x_i(t)$ the signal of microphone $i$, $y_i$ its lateral position, $\mathcal{L}$ and $\mathcal{R}$ the two halves, and $w_i$ a Gaussian taper. As the aperture of each half spans between two and a half and five wavelengths over the band of the call, the beams are wide at \SI{25}{\kilo\hertz} and narrow at \SI{50}{\kilo\hertz}, which gives the virtual ears a frequency-dependent directivity, similar to the HRTF characteristics of the real bats used in simulation. As the call covers only one octave, we used 16 pooled channels between \SI{25}{\kilo\hertz} and \SI{50}{\kilo\hertz}.

\textbf{Ground truth and odometry:} The ground truth was computed from the lidar scans with PIN-SLAM \cite{panPINSLAMLiDARSLAM2024}, and the clock offset between sonar and lidar was estimated by comparing the acoustic images with images predicted from the lidar map. As the platform recorded no wheel odometry, we derived it from the ground truth with a scale error of \SI{-5}{\percent} on distance and \SI{+10}{\percent} on rotation, plus the random noise of the simulations. On a figure-eight, the rotation error does not cancel, and at the first revisit, the heading is off by \SI{36}{\degree}. Dead reckoning has an error of \RealATEodo{}~\si{\meter}, and all results are the mean (standard deviation) over ten noise seeds.

\subsection{Results}
\label{sec:realResults}

\begin{figure*}[t]
\centering
\includegraphics[width=\linewidth]{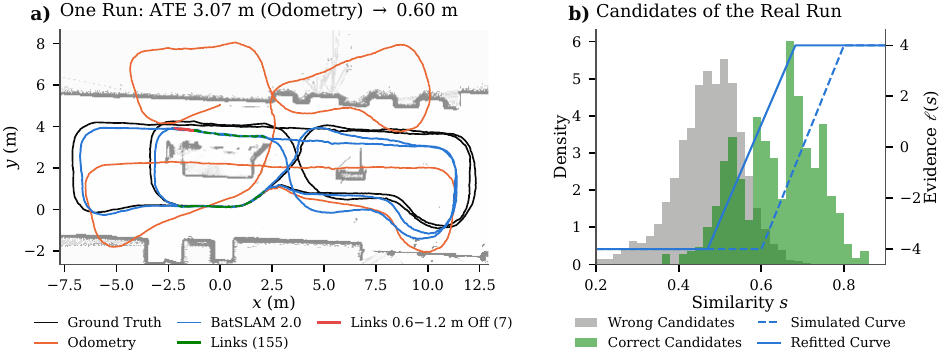}
\caption{Real-world results. a) Map of one run (refitted evidence curve) with odometry, ground truth, BatSLAM 2.0 estimate and links. b) Similarity of correct and wrong candidates, with the simulated (dashed) and refitted (solid) evidence curves.}
\label{fig:realResults}
\end{figure*}

\textbf{The evidence curve:} The real similarities $s$ (equation \ref{eq:similarity}) are lower than the simulated ones. The correct candidates had a median similarity of \RealSimTrueMed{}, and \SI{99}{\percent} of the wrong ones stayed below \RealSimFalsePnn{} (figure \ref{fig:realResults}b). Both classes are well separated, but the evidence curve of the simulations (equation \ref{eq:evidenceCurve}, with its zero crossing at \LLRmid{}) lies above most correct matches. A correct pair with the median similarity then contributes $\ell = \LLRslope \cdot (\RealSimTrueMed - \LLRmid) = -2$ to the evidence $\Lambda(h)$ of equation \ref{eq:evidence}, so that a hypothesis of typical correct pairs loses evidence with every pulse instead of gaining it, and only stretches with similarities above \LLRmid{} can reach the commit threshold. We therefore refitted the two constants of equation \ref{eq:evidenceCurve} with the same logistic regression as in the simulations (equation \ref{eq:llr}), on the candidates of one half of the drive (alternating stretches of \SI{5}{\meter}), labeled with the lidar ground truth, and ran the complete drive with it. Both folds gave similar curves (a zero crossing at \RealMidEven{} and \RealMidOdd{}). This curve is a property of the sensor, and has to be determined once per sensor rather than per deployment.

\textbf{The map:} Table \ref{tab:real} in the appendix and figure \ref{fig:realResults}a summarize the results. With either evidence curve, none of the 30 runs committed a wrong hypothesis or collapsed. With the curve of the simulations, the system linked only half of the revisits, and reduced the trajectory error from \RealATEodo{}~\si{\meter} to \RealATESim{}~\si{\meter}. With the refitted curve, it linked \RealCovEven{} of the revisits and reached \RealATEEven{}~\si{\meter} (\RealATEalEven{}~\si{\meter} after alignment). The remaining error stems mostly from the first \SI{46}{\meter} of the drive, before the first same-direction revisit. Places in the right loop were only revisited in the opposite direction, and were, as expected, never linked. With the refitted curve, a few links (\RealWrongLinksEven{} per run on average) at the tail of one correct hypothesis were \SI{0.6}{\meter} to \SI{1.2}{\meter} off, at a place where the platform turned off a path that it had previously driven straight on. The long-baseline tolerance of the verifier allows such a slow divergence.

\textbf{Odometry errors:} With a rotation error of \SI{15}{\percent} instead of \SI{10}{\percent}, the heading error at the first revisit grows to \SI{53}{\degree}, which the plausibility gate considers implausible given the heading noise assumed by the back-end. The correct loop closures are then rejected (\RealATEcliffEven{}~\si{\meter}), but no wrong loop closure is committed either, and with twice the assumed heading noise, the system recovers (\RealATEcliffNoise{}~\si{\meter}). Finally, the system runs in real time on this drive, with \RealFrontendMs{}~\si{\milli\second} for the front-end and \RealSlamMs{}~\si{\milli\second} for the rest of the system per pulse on average, against \RealBudgetMs{}~\si{\milli\second} between two calls.

\section{Discussion and limitations}
\label{sec:limitations}

In contrast to SeqSLAM \cite{milfordSeqSLAMVisualRoutebased2012}, temporal consistency alone is insufficient for sonar: a short segment of one corridor can genuinely produce an echo sequence that closely resembles a segment elsewhere. The main remaining vulnerability therefore occurs early in a drive, when the relative pose uncertainty between two places may still span several meters and a strong aliased match remains geometrically plausible. Postponing all large corrections would reduce this risk, but would also delay the loop closures needed to constrain the map. Our ablations expose this trade-off directly: requiring 24 supporting pairs eliminated the remaining alias for the \SI{-20}{\deci\bel} sensor, but left the map of route 7 rotated. For a purely topological representation, the stricter criterion is therefore preferable, as the global rotation of the map is not that important for practical applications. For metric SLAM, however, timely anchoring of the map is also important, and a tradeoff can be made here.

A second limitation is that most experiments were conducted in simulation. The simulator models sensor directivity, geometric spreading, atmospheric absorption, and reflectivity, but does not include weak higher-order reflections. The real-world experiment demonstrates that the approach transfers to an actual acoustic sensor, requiring just the evidence curve to be recalibrated. 

The current pose graph assumes that odometry errors are predominantly random. Under a systematic heading bias, the inferred topology remains correct, but metric accuracy progressively deteriorates as expected. Introducing an explicit heading-bias state into the factor graph would provide a natural way to estimate and compensate for such systematic errors. Similarly, locations revisited only in the opposite direction are currently never associated. This limitation could be addressed by explicitly modeling how a place is expected to appear acoustically when traversed from the reverse direction, but requires complex learned world models, which is part of our future work.

Finally, although the simulated environments were deliberately designed to contain substantial self-similarity, the sequence verifier rejected nearly all incorrect hypotheses before they reached the back-end. Consequently, the GNC audit did not alter any final map in the experiments reported here. Its role is likely to become more important in environments with stronger structural aliasing, such as office floors containing repeated rooms or parking garages with highly repetitive geometry, where several individually plausible but mutually inconsistent loop-closure hypotheses may coexist. However, the target application for an algorithm like BatSLAM is in environments where optical techniques tend to fail, such as in heavy industry applications (mining, agriculture, etc). In these environments, the type of degenerative aliassing as encountered in long office corridors is typically not encountered, making this issue less pressing. In addition, all parameters were fixed using a single development trajectory, whereas the real sensor required different evidence constants. These constants primarily determine loop-closure coverage rather than map safety, but currently need to be calibrated once for each sensor configuration. Self-learning using active inference of these parameters might be an interesting avenue for further research into this architecture.

\section{Conclusion}
\label{sec:conclusion}

In this paper, we presented BatSLAM 2.0, a sonar-only SLAM system that combines a biomimetic acoustic front-end with sequence-verified place recognition and a robust pose-graph back-end. The central challenge is the strong perceptual aliasing inherent to sonar. BatSLAM 2.0 addresses this ambiguity through an explicit direction cue in the place descriptor, a sequence verifier that accepts only long, strong, unambiguous, and geometrically plausible hypotheses, a commit rule that increases the required evidence with the magnitude of the implied correction, and a back-end in which each accepted loop closure remains represented as a removable group of weak constraints.

Across all \NdesRuns{} simulated runs with calibrated or approximately calibrated odometry, the system produced consistent maps without incorrect loop closures, achieving \ATEalignedDesRange{} after alignment. When systematic odometry bias was increased beyond \SI{0.3}{\degree\per\meter}, performance degraded primarily in metric accuracy rather than in topological correctness (ie, no map collapses occured). Our analysis further showed that sensor signal-to-noise ratio and spectral-shape information are central to reliable place recognition, and that the effective capture region of a stored template is limited to approximately \CaptureLateral{} laterally and \CaptureHeading{} in heading. On a real acoustic recording, BatSLAM 2.0 likewise produced a consistent map without incorrect loop closures while operating in real time, reducing the trajectory error from \RealATEodo{}~\si{\meter} to \RealATEEven{}~\si{\meter}.

Future work will focus on substantially longer real-world trajectories using physical odometry and a bat-like binaural sensing head, explicit estimation of systematic odometry errors, and place recognition across opposite traversal directions.

\section*{Data and code availability}

The BatSLAM 2.0 source code, simulator, and scripts used to generate all experimental results, tables, and figures in this paper are available at \url{https://github.com/Cosys-Lab/BatSLAM2}.

\section*{Acknowledgment}

Claude Opus 5.5 (Anthropic) was used to assist with writing the code, generating the test harness, and drafting the text of this paper. The authors substantially revised the text and take full responsibility for the content of this paper.

\bibliographystyle{IEEEtran}
\bibliography{references}

\clearpage
\onecolumn
\appendices
\section{Parameters}
\label{app:params}
Table \ref{tab:params} lists all parameters of BatSLAM 2.0. They are the default values of \texttt{batslam/config.py} in the accompanying code, and were used for all worlds, routes and sensors; none of them was tuned per test case.

\begin{table*}[t]
\centering
\footnotesize
\caption{Parameters of BatSLAM 2.0. The same values were used for every experiment in this paper.}
\label{tab:params}
\resizebox{\textwidth}{!}{\begin{tabular}{lll}
\toprule
Stage & Parameter & Value \\
\midrule
Front-end & Filterbank & 64 Gaussian bands, \SIrange{30}{90}{\kilo\hertz}, log-spaced, width of two band spacings \\
 & Band pooling & Pairs of bands (32 bands) \\
 & Range bins & \SI{6}{\centi\meter}, from \SI{0.25}{\meter} to \SI{10}{\meter} (162 bins) \\
 & Time-varying gain & Amplitude $\times r$ \\
 & Energy image & Cube root of the amplitude, relative to the peak of the view \\
 & Spectral-shape image & $\pm$\SI{15}{\deci\bel} scaled to $\pm 1$, only above \SI{-40}{\deci\bel} \\
 & Range smoothing & Gaussian, one bin (\SI{6}{\centi\meter}) \\
Templates & Spacing & \SI{0.45}{\meter} travelled or \SI{20}{\degree} turned \\
 & Range-shift search & $\pm 3$ bins ($\pm$\SI{18}{\centi\meter}) \\
 & Candidates & 5 per pulse, similarity $\ge \Smin$ \\
 & Exclusion of recent templates & \SI{6}{\meter} of travel \\
 & Heading gate & \SI{69}{\degree} (\SI{1.2}{\radian}) \\
Verifier & Evidence per match & $\LLRslope \cdot (s - \LLRmid)$, clipped to $\pm 4$ \\
 & Missed pulse & Penalty 0.5, dies after 4 missed pulses \\
 & Consistency & \SI{0.30}{\meter} $+ 0.15 \times$ distance, \SI{0.25}{\radian} \\
 & Commit & 8 pairs, 3 templates, evidence 10, margin 4 over rivals ($>$\SI{1}{\meter} or $>$\SI{0.4}{\radian} apart) \\
 & Plausibility gate & $\chi^2_3$ at 99.9\,\% (16.27); relative covariance for corrections $>$\SI{1.5}{\meter} \\
 & Risk-scaled commit & Correction $>$\SI{1.5}{\meter}: 16 pairs, evidence 40, evidence $\ge$ number of pairs \\
Pose graph & Prior & $\sigma = (0.01, 0.01, 0.005)$ \\
 & Odometry & $\sigma = (0.04, 0.02, 0.035) \cdot \sqrt{d} + (0.005, 0.005, 0.003)$ \\
 & Loop closure link & $\sigma = (0.30, 0.30, 0.20)$, Huber loss with threshold 1.345 \\
 & Solver & iSAM2, Dogleg, relinearization threshold 0.01 \\
Link management & Length rule & Fewer than 16 links after 10 pulses without a new link \\
 & Residual check & Median $\chi^2 > 12$ (tentative groups) \\
 & GNC audit & Every 400 nodes, Geman-McClure loss, keep if mean weight $\ge 0.5$ \\
\bottomrule
\end{tabular}}
\end{table*}

\section{Reproducibility}
\label{app:repro}
All code, scene definitions and experiment definitions are available at \url{https://github.com/Cosys-Lab/BatSLAM2}. The simulator (section \ref{sec:setup}) is included as a copy under \texttt{sim/simulator}. An important note concerns the HRTF data: the HRTF file that we started from stores its directivity arrays in (elevation, azimuth, frequency) order, while the interpolation routine of the simulator assumed (azimuth, elevation, frequency). As both angle grids are identical ($-$\SI{90}{\degree} to \SI{90}{\degree} in steps of \SI{2.5}{\degree}), this went unnoticed at first, and it caused the simulated ears to point up and down instead of left and right (a reflector \SI{45}{\degree} to the side produced an interaural level difference of only \SI{0.2}{\deci\bel}, and a reflector \SI{30}{\degree} above the head one of \SI{14}{\deci\bel}). All results in this paper were produced with the corrected (transposed) HRTF, for which the same test gives \SI{18}{\deci\bel} and \SI{0}{\deci\bel}, respectively. The check is included as \texttt{sim/check\_hrtf\_orientation.m}.

The datasets are generated with \texttt{sim/generate\_dataset.m} (worlds, drives, echoes and odometry) and \texttt{sim/generate\_capture\_dataset.m} (the capture study). The experiments are defined in \texttt{suites/hc\_main.json}, \texttt{suites/hc\_ablation.json} and \texttt{suites/hc\_robustness.json}, and are run with \texttt{scripts/run\_suite.py}. The front-end studies are run with \texttt{scripts/bench\_descriptor.py}, \texttt{scripts/bench\_sensor.py} and \texttt{scripts/capture\_analysis.py}. All numbers and tables in this paper are generated from the results by \texttt{scripts/build\_paper\_generated.py}, and all figures by \texttt{scripts/paper\_figures.py}.

\section{Result tables}
\label{app:tables}
This appendix lists the numerical results that are discussed in the main text: single-view place recognition for the design choices of the descriptor (table \ref{tab:descriptor}), the results of the complete system per condition (table \ref{tab:main}), the ablations (table \ref{tab:ablation}) and the real-world results (table \ref{tab:real}).

\begin{table}[t]
\centering
\footnotesize
\caption{Single-view place recognition for the design choices of the descriptor, with the design sensor. All variants use 32 channels, \SI{6}{\centi\meter} range bins and the TVG. Our descriptor uses a cube-root compression. Best value per column in bold. ILD: interaural level difference.}
\label{tab:descriptor}
\begin{tabular}{lrrrr}
\toprule
 & \multicolumn{2}{c}{Indoor floor} & \multicolumn{2}{c}{City} \\
\cmidrule(lr){2-3}\cmidrule(lr){4-5}
Variant & Top-1 & AUC & Top-1 & AUC \\
\midrule
\multicolumn{5}{l}{\textit{Cues}} \\
Energy & 0.875 & 0.915 & 0.878 & 0.918 \\
Energy + shape \textbf{(ours)} & 0.913 & 0.945 & 0.888 & 0.938 \\
Energy + shape + ILD & 0.892 & \textbf{0.968} & 0.878 & \textbf{0.960} \\
\midrule
\multicolumn{5}{l}{\textit{Compression of the energy image (energy + shape)}} \\
Logarithm, floor \SI{-30}{\deci\bel} & 0.884 & 0.930 & 0.888 & 0.930 \\
Logarithm, floor \SI{-50}{\deci\bel} & \textbf{0.929} & 0.947 & 0.886 & 0.925 \\
Square root & 0.900 & 0.940 & \textbf{0.889} & 0.937 \\
Linear & 0.858 & 0.925 & 0.857 & 0.921 \\
\midrule
\multicolumn{5}{l}{\textit{Matching of the original BatSLAM}} \\
Logarithmic energy, floor \SI{-30}{\deci\bel} & 0.401 & 0.801 & 0.599 & 0.814 \\
Energy + shape & 0.793 & 0.901 & 0.802 & 0.889 \\
\bottomrule
\end{tabular}
\end{table}

\begin{table}[t]
\centering
\footnotesize
\caption{Results of BatSLAM 2.0 per condition (ranges over the runs). ATE: trajectory error of odometry and of BatSLAM 2.0, without and after rigid alignment; wrong: wrong committed hypotheses in the final graph, out of all committed hypotheses; coverage: fraction of true revisits with a correct link.}
\label{tab:main}
\begin{tabular}{lrrrrr}
\toprule
 & \multicolumn{3}{c}{ATE [m]} & & \\
\cmidrule(lr){2-4}
Condition (runs) & Odometry & Ours & Aligned & Wrong & Coverage \\
\midrule
Calibrated (6) & 4.2--11.1 & 0.47--1.91 & 0.18--0.46 & 0/230 & 0.79--0.84 \\
Residual bias (6) & 11.0--19.0 & 0.62--1.65 & 0.23--0.53 & 0/227 & 0.79--0.84 \\
Degraded sensors (3) & 7.5 & 0.52--1.79 & 0.25--0.81 & 1/106 & 0.67--0.74 \\
Uncalibrated (4) & 19.1--22.3 & 1.38--12.68 & 0.45--2.96 & 0/149 & 0.62--0.81 \\
\bottomrule
\end{tabular}
\end{table}

\begin{table}[t]
\centering
\footnotesize
\caption{Ablations over four standard cases (routes 3 and 5 with calibrated odometry, route 7 with a residual bias, and the city). ATE: mean over the cases; wrong: wrong committed hypotheses in the final graphs, summed over the cases. Naive: no sequence verification and no link management.}
\label{tab:ablation}
\setlength{\tabcolsep}{5pt}
\begin{tabular}{lrrr}
\toprule
Variant & ATE [m] & Wrong & Coverage \\
\midrule
\textbf{Full system (ours)} & 0.87 & 0 & 0.81 \\
\midrule
Naive & 18.90 & 364 & 0.72 \\
No sequence verification & 4.85 & 10 & 0.80 \\
\midrule
Original front-end (8 bands) & 1.54 & 0 & 0.73 \\
No time-varying gain & 2.53 & 3 & 0.72 \\
Energy image only & 1.38 & 8 & 0.74 \\
No range-shift search & 1.47 & 0 & 0.60 \\
Hand-set evidence constants & 1.51 & 0 & 0.68 \\
\midrule
No risk-scaled commit & 0.93 & 1 & 0.81 \\
Length only (24 pairs) & 1.20 & 0 & 0.81 \\
No link management & 0.55 & 0 & 0.82 \\
\bottomrule
\end{tabular}
\end{table}

\begin{table*}[t]
\centering
\caption{Real-world results, mean (standard deviation) over ten odometry seeds (distance $\times 0.95$, rotation $\times 1.10$).}
\label{tab:real}
\begin{tabular}{lrrrrrrr}
\toprule
Evidence curve & ATE & Aligned & Links & Precision & Wrong & Revisits & Collapsed \\
 & (\si{\meter}) & (\si{\meter}) & & & Hyp. & Linked & Pairs \\
\midrule
Odometry only & 3.62~(0.46) & & & & & & \\
Simulation evidence curve & 0.77~(0.07) & 0.60~(0.05) & 108 & 1.000~(0.000) & 0 & 0.50~(0.01) & 0.000 \\
Refitted, even stretches & 0.67~(0.13) & 0.52~(0.11) & 141 & 0.980~(0.018) & 0 & 0.64~(0.10) & 0.000 \\
Refitted, odd stretches & 0.66~(0.12) & 0.52~(0.11) & 142 & 0.975~(0.021) & 0 & 0.64~(0.10) & 0.000 \\
\bottomrule
\end{tabular}

\end{table*}

\clearpage
\section{Additional results}
\label{app:runs}
In this appendix, we show the trajectories and the detailed results of all runs of the main and robustness experiments, and of the ablations. In every map, the grey line is the ground truth, the orange line the dead-reckoning estimate, and the blue line the trajectory estimated by BatSLAM 2.0 (without alignment, as it was estimated online). Links of wrong committed hypotheses that remain in the final graph would be drawn in red. The title of every panel lists the trajectory error of BatSLAM 2.0 and of odometry, and the number of wrong committed hypotheses in the final graph.

\begin{figure*}[p]
\centering
\includegraphics[width=\linewidth]{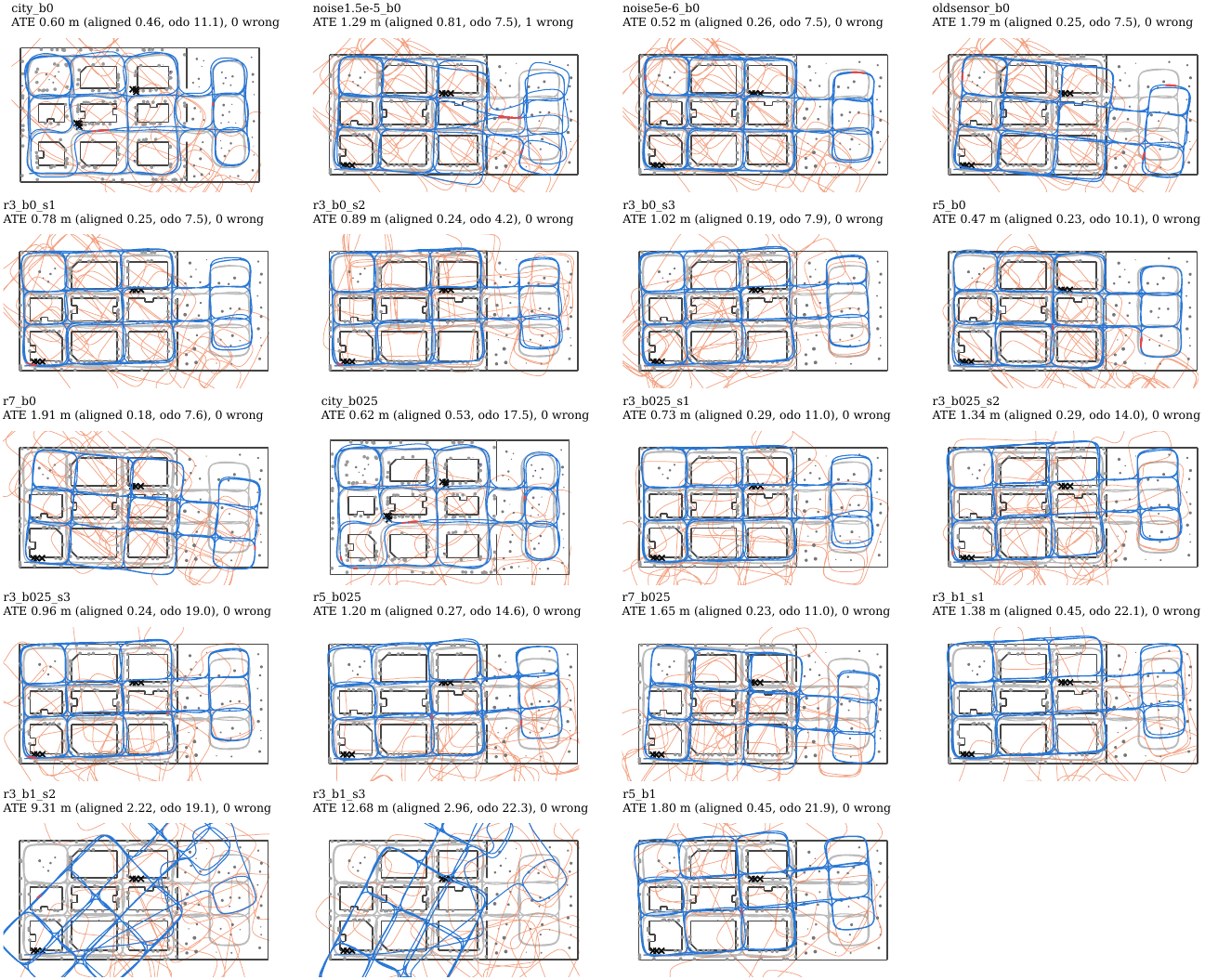}
\caption{Trajectories of all runs of the main experiment (table \ref{tab:mainRuns}), sorted by the odometry condition. Grey: ground truth; orange: odometry; blue: BatSLAM 2.0. Run names: \texttt{r3}, \texttt{r5}, \texttt{r7}: routes 3, 5 and 7 on the indoor floor; \texttt{city}: the city world; \texttt{b0}, \texttt{b025}, \texttt{b1}: bias factor $b = 0$, $0.25$ and $1$; \texttt{s1}--\texttt{s3}: odometry noise realization; \texttt{noise5e-6}, \texttt{noise1.5e-5}, \texttt{oldsensor}: degraded sensors.}
\label{fig:appendixMaps}
\end{figure*}

\begin{figure*}[p]
\centering
\includegraphics[width=\linewidth]{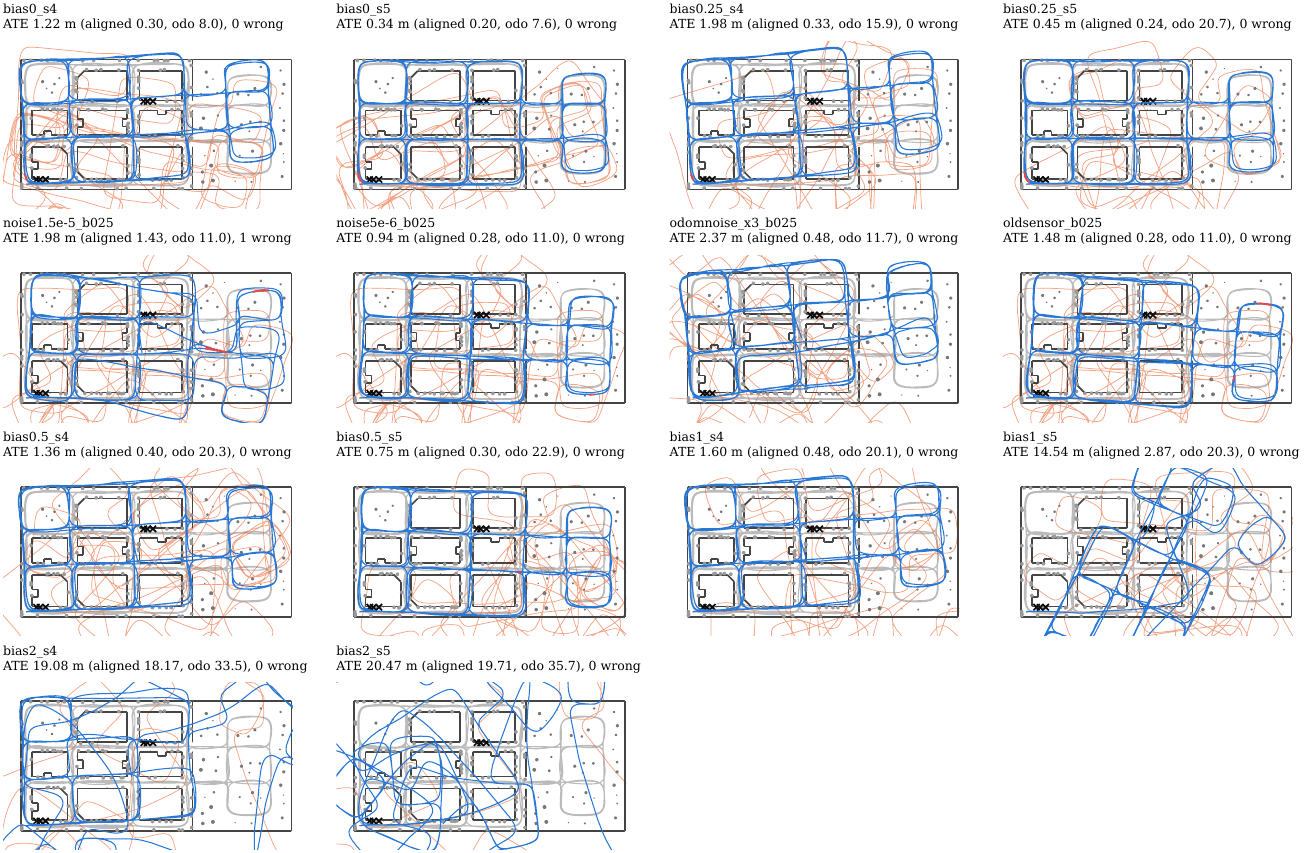}
\caption{Trajectories of all runs of the robustness experiment (table \ref{tab:robustness}). \texttt{bias}$b$\texttt{\_s}$k$: bias factor $b$ with odometry realization $k$; \texttt{odomnoise\_x3}: three times the random odometry noise; the last three runs use degraded sensors with a residual bias. Colors as in figure \ref{fig:appendixMaps}.}
\label{fig:appendixMapsRobust}
\end{figure*}

\begin{table*}[htbp]
\centering
\footnotesize
\caption{Results of BatSLAM 2.0 on all runs of the main experiment (drives of about \SI{1}{\kilo\meter}). ATE odo: dead reckoning; ATE and aligned: BatSLAM 2.0 without and after rigid alignment; precision: fraction of correct links; wrong (final / all): wrong committed hypotheses in the final graph and over all commits; coverage: fraction of true revisits with a correct link; collapsed: fraction of collapsed pairs.}
\label{tab:mainRuns}
\resizebox{\linewidth}{!}{\begin{tabular}{llrrrrrrrr}
\toprule
Case & Odometry & ATE odo & ATE & Aligned & Precision & Wrong (final) & Wrong (all) & Coverage & Collapsed \\
 & & [m] $\downarrow$ & [m] $\downarrow$ & [m] $\downarrow$ & $\uparrow$ & $\downarrow$ & $\downarrow$ & $\uparrow$ & [\%] $\downarrow$ \\
\midrule
Indoor, route 3, seed 1 & Calibrated & 7.5 & \textbf{0.78} & 0.25 & 0.999 & 0/39 & 0 & 0.84 & 0.0 \\
Indoor, route 3, seed 2 & Calibrated & 4.2 & \textbf{0.89} & 0.24 & 0.999 & 0/40 & 0 & 0.84 & 0.0 \\
Indoor, route 3, seed 3 & Calibrated & 7.9 & \textbf{1.02} & 0.19 & 1.000 & 0/36 & 1 & 0.84 & 0.0 \\
Indoor, route 5 & Calibrated & 10.1 & \textbf{0.47} & 0.23 & 0.998 & 0/45 & 0 & 0.82 & 0.0 \\
Indoor, route 7 & Calibrated & 7.6 & \textbf{1.91} & 0.18 & 1.000 & 0/46 & 0 & 0.80 & 0.0 \\
City & Calibrated & 11.1 & \textbf{0.60} & 0.46 & 0.996 & 0/24 & 0 & 0.79 & 0.0 \\
Indoor, route 3, sensor $-10$~dB & Calibrated & 7.5 & \textbf{0.52} & 0.26 & 0.997 & 0/34 & 0 & 0.72 & 0.0 \\
Indoor, route 3, sensor $-20$~dB & Calibrated & 7.5 & \textbf{1.29} & 0.81 & 0.989 & 1/32 & 2 & 0.67 & 0.1 \\
Indoor, route 3, original sensor & Calibrated & 7.5 & \textbf{1.79} & 0.25 & 0.995 & 0/40 & 3 & 0.74 & 0.0 \\
\midrule
Indoor, route 3, seed 1 & Residual bias & 11.0 & \textbf{0.73} & 0.29 & 1.000 & 0/38 & 1 & 0.83 & 0.0 \\
Indoor, route 3, seed 2 & Residual bias & 14.0 & \textbf{1.34} & 0.29 & 1.000 & 0/38 & 0 & 0.84 & 0.0 \\
Indoor, route 3, seed 3 & Residual bias & 19.0 & \textbf{0.96} & 0.24 & 0.999 & 0/40 & 1 & 0.84 & 0.0 \\
Indoor, route 5 & Residual bias & 14.6 & \textbf{1.20} & 0.27 & 0.999 & 0/41 & 0 & 0.80 & 0.0 \\
Indoor, route 7 & Residual bias & 11.0 & \textbf{1.65} & 0.23 & 1.000 & 0/42 & 0 & 0.79 & 0.0 \\
City & Residual bias & 17.5 & \textbf{0.62} & 0.53 & 0.994 & 0/28 & 0 & 0.82 & 0.0 \\
\midrule
Indoor, route 3, seed 1 & Uncalibrated & 22.1 & \textbf{1.38} & 0.45 & 1.000 & 0/37 & 1 & 0.81 & 0.0 \\
Indoor, route 3, seed 2 & Uncalibrated & 19.1 & \textbf{9.31} & 2.22 & 0.998 & 0/36 & 1 & 0.62 & 0.6 \\
Indoor, route 3, seed 3 & Uncalibrated & 22.3 & \textbf{12.68} & 2.96 & 0.999 & 0/33 & 1 & 0.62 & 0.8 \\
Indoor, route 5 & Uncalibrated & 21.9 & \textbf{1.80} & 0.45 & 1.000 & 0/43 & 0 & 0.75 & 0.0 \\
\bottomrule
\end{tabular}
}
\end{table*}

\begin{table*}[htbp]
\centering
\footnotesize
\caption{Robustness runs: odometry bias factor $b$, three times the random odometry noise, and degraded sensors with a residual bias. Columns as in table \ref{tab:mainRuns}.}
\label{tab:robustness}
\resizebox{\linewidth}{!}{\begin{tabular}{llrrrrrrrr}
\toprule
Case & Bias & ATE odo & ATE & Aligned & Precision & Wrong (final) & Wrong (all) & Coverage & Collapsed \\
 & & [m] $\downarrow$ & [m] $\downarrow$ & [m] $\downarrow$ & $\uparrow$ & $\downarrow$ & $\downarrow$ & $\uparrow$ & [\%] $\downarrow$ \\
\midrule
Indoor, route 3, seed 4 & $b = 0$ & 8.0 & \textbf{1.22} & 0.30 & 1.000 & 0/38 & 1 & 0.82 & 0.0 \\
Indoor, route 3, seed 5 & $b = 0$ & 7.6 & \textbf{0.34} & 0.20 & 0.999 & 0/40 & 1 & 0.83 & 0.0 \\
Indoor, route 3, seed 4 & $b = 0.25$ & 15.9 & \textbf{1.98} & 0.33 & 0.999 & 0/38 & 1 & 0.83 & 0.0 \\
Indoor, route 3, seed 5 & $b = 0.25$ & 20.7 & \textbf{0.45} & 0.24 & 0.999 & 0/37 & 2 & 0.83 & 0.0 \\
Indoor, route 3, sensor $-20$~dB & $b = 0.25$ & 11.0 & \textbf{1.98} & 1.43 & 0.988 & 1/33 & 3 & 0.65 & 0.2 \\
Indoor, route 3, sensor $-10$~dB & $b = 0.25$ & 11.0 & \textbf{0.94} & 0.28 & 1.000 & 0/37 & 2 & 0.76 & 0.0 \\
Indoor, route 3, seed 1, odometry noise $\times 3$ & $b = 0.25$ & 11.7 & \textbf{2.37} & 0.48 & 1.000 & 0/50 & 1 & 0.81 & 0.0 \\
Indoor, route 3, original sensor & $b = 0.25$ & 11.0 & \textbf{1.48} & 0.28 & 0.998 & 0/38 & 2 & 0.73 & 0.0 \\
Indoor, route 3, seed 4 & $b = 0.5$ & 20.3 & \textbf{1.36} & 0.40 & 1.000 & 0/38 & 1 & 0.82 & 0.0 \\
Indoor, route 3, seed 5 & $b = 0.5$ & 22.9 & \textbf{0.75} & 0.30 & 1.000 & 0/40 & 1 & 0.82 & 0.0 \\
Indoor, route 3, seed 4 & $b = 1$ & 20.1 & \textbf{1.60} & 0.48 & 1.000 & 0/40 & 1 & 0.83 & 0.0 \\
Indoor, route 3, seed 5 & $b = 1$ & 20.3 & \textbf{14.54} & 2.87 & 1.000 & 0/32 & 1 & 0.62 & 0.8 \\
Indoor, route 3, seed 4 & $b = 2$ & 33.5 & \textbf{19.08} & 18.17 & 1.000 & 0/23 & 0 & 0.41 & 4.8 \\
Indoor, route 3, seed 5 & $b = 2$ & 35.7 & \textbf{20.47} & 19.71 & 1.000 & 0/11 & 0 & 0.32 & 7.4 \\
\bottomrule
\end{tabular}
}
\end{table*}

\begin{table*}[htbp]
\centering
\footnotesize
\caption{Ablation of the safeguards on the four hardest cases (three degraded sensors and route 3 with odometry seed 3). Every cell lists the aligned trajectory error (m) and, in parentheses, the number of wrong hypotheses in the final graph; the last two columns sum the wrong hypotheses in the final graphs and over all commits.}
\label{tab:ablationHard}
\resizebox{\linewidth}{!}{\begin{tabular}{lrrrrrr}
\toprule
Variant & $-10$~dB sensor & $-20$~dB sensor & Original sensor & Route 3, seed 3 & Wrong (final) & Wrong (all) \\
\midrule
No risk-scaled commit & 0.26 (0) & 0.24 (0) & 1.06 (1) & 0.19 (0) & 1 & 10 \\
Risk-scaled commit, length only (24 pairs) & 0.26 (0) & 0.24 (0) & 0.25 (0) & 0.19 (0) & 0 & 5 \\
Plausibility gate on the absolute marginal & 0.26 (0) & 0.23 (0) & 0.69 (1) & 0.19 (0) & 1 & 12 \\
No plausibility gate & 0.26 (0) & 1.37 (2) & 0.60 (1) & 0.19 (0) & 3 & 19 \\
No length rule & 0.26 (0) & 0.81 (2) & 1.07 (3) & 0.20 (2) & 7 & 7 \\
No link management & 0.26 (0) & 0.81 (2) & 1.07 (3) & 0.20 (2) & 7 & 7 \\
\midrule
\textbf{Full system (ours)} & 0.26 (0) & 0.81 (1) & 0.25 (0) & 0.19 (0) & 1 & 6 \\
\bottomrule
\end{tabular}
}
\end{table*}

\begin{table*}[htbp]
\centering
\footnotesize
\caption{Ablations per case: trajectory error (ATE, in m) and the number of wrong committed hypotheses in the final graph (in parentheses), for route 3 and route 5 with calibrated odometry, route 7 with a residual bias, and the city with calibrated odometry.}
\label{tab:ablationCases}
\begin{tabular}{lrrrr}
\toprule
Variant & Route 3 & Route 5 & Route 7, residual bias & City \\
\midrule
Naive (single views, no gates, no link mgmt.) & 8.00 (63) & 26.70 (68) & 8.22 (72) & 32.68 (161) \\
No sequence verification (single views) & 5.08 (1) & 1.95 (1) & 8.17 (6) & 4.21 (2) \\
Original front-end (8 wide bands) & 1.97 (0) & 0.44 (0) & 3.15 (0) & 0.59 (0) \\
No time-varying gain & 0.85 (1) & 6.82 (2) & 1.84 (0) & 0.62 (0) \\
Energy image only (no shape) & 0.63 (0) & 1.70 (3) & 1.26 (4) & 1.92 (1) \\
No range-shift search & 0.45 (0) & 1.57 (0) & 3.28 (0) & 0.56 (0) \\
No risk-scaled commit & 0.88 (0) & 0.46 (0) & 1.79 (1) & 0.57 (0) \\
Risk-scaled commit, length only (24 pairs) & 0.74 (0) & 0.47 (0) & 2.98 (0) & 0.60 (0) \\
Risk-scaled commit, length only (16 pairs) & 0.78 (0) & 0.47 (0) & 1.78 (1) & 0.52 (0) \\
No length rule & 0.63 (0) & 0.47 (0) & 0.49 (0) & 0.63 (0) \\
No link management & 0.63 (0) & 0.47 (0) & 0.49 (0) & 0.63 (0) \\
No plausibility gate & 0.78 (0) & 0.47 (0) & 1.65 (0) & 0.60 (0) \\
Plausibility gate on the absolute marginal & 0.78 (0) & 0.47 (0) & 1.65 (0) & 0.60 (0) \\
No heading gate & 0.64 (0) & 0.42 (0) & 1.55 (0) & 0.50 (0) \\
Hand-set evidence constants & 0.68 (0) & 0.63 (0) & 3.29 (0) & 1.45 (0) \\
\midrule
\textbf{Full system (ours)} & 0.78 (0) & 0.47 (0) & 1.65 (0) & 0.60 (0) \\
\bottomrule
\end{tabular}

\end{table*}

\end{document}